\documentclass{article}
\usepackage{url}
\usepackage{PRIMEarxiv}
\usepackage[utf8]{inputenc} 
\usepackage[T1]{fontenc}    
\usepackage{hyperref}       
\usepackage{url}            
\usepackage{booktabs}       
\usepackage{amsfonts}       
\usepackage{nicefrac}       
\usepackage{microtype}      
\usepackage{lipsum}
\usepackage{fancyhdr}       
\usepackage{graphicx}       
\graphicspath{{media/}}     
\usepackage{multirow}
\usepackage{graphicx}
\usepackage{amsmath,amssymb,amsfonts}
\usepackage{smartdiagram}  
\usepackage{booktabs}
\usepackage{textcomp}
\usepackage{ctable}
\usepackage{xcolor}
\usepackage[symbol]{footmisc}
\usepackage{authblk} 
\hypersetup{
  colorlinks=True,
  citecolor=cyan,
  linkcolor=cyan,
  urlcolor=cyan}

\title{A Novel real-time arrhythmia detection model using YOLOv8}

\begin{document}
\author[a,1]{\textbf{Guang Jun Nicholas Ang}} 
\author[a,1]{\textbf{Aritejh Kr Goil}}
\author[a,b]{\textbf{Henryk Chan}}
\author[a]{\textbf{Jieyi Jeric Lew}}
\author[a]{\textbf{Xin Chun Lee}}
\author[a]{\textbf{Raihan Bin Ahmad Mustaffa}}
\author[a]{\textbf{Timotius Jason}}
\author[a]{\textbf{Ze Ting Woon}}
\author[a,c]{\textbf{Bingquan Shen}}

\affil[a]{National University of Singapore, Singapore}
\affil[b]{University of Sheffield, United Kingdom}
\affil[c]{DSO National Laboratories, Singapore}
\affil[1]{\{anggjnicholas, aritejh\}@u.nus.edu}

\maketitle

\begin{abstract}
In a landscape characterized by heightened connectivity and mobility, coupled with a surge in cardiovascular ailments, the imperative to curtail healthcare expenses through remote monitoring of cardiovascular health has become more pronounced. The accurate detection and classification of cardiac arrhythmias are pivotal for diagnosing individuals with heart irregularities. This study underscores the feasibility of employing electrocardiograms (ECG) measurements in the home environment for real-time arrhythmia detection.

Presenting a fresh application for arrhythmia detection, this paper leverages the cutting-edge You-Only-Look-Once (YOLO)v8 algorithm to categorize single-lead ECG signals. We introduce a novel loss-modified YOLOv8 model, fine-tuned on the MIT-BIH arrhythmia dataset, enabling real-time continuous monitoring. The obtained results substantiate the efficacy of our approach, with the model attaining an average accuracy of 99.5\% and 0.992 mAP@50, and a rapid detection time of 0.002 seconds on an NVIDIA Tesla V100.

Our investigation exemplifies the potential of real-time arrhythmia detection, enabling users to visually interpret the model output within the comfort of their homes. Furthermore, this study lays the groundwork for an extension into a real-time explainable AI (XAI) model capable of deployment in the healthcare sector, thereby significantly advancing the realm of healthcare solutions.
\end{abstract}


\section{Introduction} \label{introduction}
Cardiac arrhythmia (or heart rhythm disorders) is a ubiquitous global ailment that occurs in 2.35\% of adults, accounting for 60\% of deaths caused by cardiovascular disease \cite{cdc_database, hammad2021automated, grabowski2022classification}. Cardiac arrhythmia is broadly categorised by the origin of the abnormally occurring beat, medically known as an ectopic beat \cite{DOWD2014}. Supraventricular ectopic beats are arrhythmias with sinus, atrial, or nodal origins, typically associated with abnormalities in the P wave. Ventricular ectopic beats are arrhythmias with ventricular origins, typically associated with abnormalities in the QRS complex. The associated waves' rate, regularity, presence, absence, and morphology further classify specific arrhythmias within the category. Arrhythmias with atrial origin include atrial fibrillation, where P waves are absent, and atrial flutter, where P waves have a distinctive sawtooth appearance. Ventricular tachycardia is a ventricular ectopic beat where the QRS complex is distinctively wide, indicating a conduction delay in the ventricles, which may cause hypotension \cite{pham2021novel,yildirim2018arrhythmia, martis2013application}. Most arrhythmia symptoms are non-fatal, including palpitations, dizziness, shortness of breath and fainting. However, if left untreated for prolonged periods, arrhythmia may become life-threatening, manifesting as heart failure and hypotension \cite{hammad2021automated, mathews2018novel}. In addition, cardiac arrhythmias may be intermittent, making it difficult for in-clinic evaluations. Hence, long-term arrhythmia monitoring is crucial, as seen from the growing number of innovations and investigations for daily monitoring \cite{liu2022multiclass}. Thus, early detection and classification of electrocardiogram (ECG) signals are crucial to diagnosing and treating arrhythmia to prevent the onset of life-threatening conditions \cite{luz2016ecg}. The ECG is a non-invasive test that reflects the vector sum of the action potential in the heart throughout successive cardiac cycles as deflections on a voltage versus time graph \cite{feather2020kumar, lilly2012pathophysiology, grabowski2022classification}. For example, on the ECG, the P wave is the vector sum of the action potential during atrial depolarisation. Likewise, the QRS complex reflects the action potential associated with the sequential depolarisation of the septum, ventricles, and ventricular myocardium. Finally, the T wave occurs in conjunction with the ventricular myocardium's re-polarisation or relaxation. By analysing the temporal and morphological features of the waves, cardiologists refer to 12-lead ECGs to identify deviations in the cardiac rhythm to make a diagnosis \cite{bigler2021accuracy, prabhakararao2020myocardial, lyakhov2021system}. Interpretation of the ECG highlights structural and functional abnormalities of the heart to aid in diagnosing cardiovascular diseases \cite{glass2006cardiac}. However, interpretation requires expert knowledge and continuous monitoring over an extended period through ambulatory Holter devices \cite{pham2021novel, loh2022application}. Furthermore, since Holter devices do not analyse the ECG, the high signal data makes manual interpretation time-consuming and prone to fatigue-induced error.  

Machine learning has led to the development of computer-aided diagnostic systems (CADS) that can automatically classify ECGs to assist cardiologists in detecting arrhythmia from long-term ECG recordings. These systems employ feature engineering techniques such as Hermite functions and polynomials, wavelet-based features, and ECG morphology to extract features from ECG signals \cite{hermitefunc, wavelettransform, hu1997patient}. Subsequently, machine learning models, including the k-th nearest-neighbours (KNN) algorithm, decision trees, and support vector machines (SVMs), are used to match these complex features to represent the preprocessed ECG signal as a sequence of stochastic patterns \cite{svm1,knn1, knn2, decision3}. Unfortunately, the combined use of feature engineering and dimensionality reduction algorithms significantly increased the computational complexity of the overall process, thereby limiting the usage of portable or wearable health monitoring devices \cite{luz2016ecg}. In recent years, deep learning (DL) has addressed the computational complexity of machine learning frameworks, using artificial neural networks with multiple hidden layers to automatically learn complex and non-linear relationships \cite{lecun}. Various DL methods, including convolutional neural networks (CNN), recurrent neural networks (RNN), and long short-term memory (LSTM)  \cite{pham2021novel,oh2018automated, yildirim2018arrhythmia, acharya2017deep}, have been applied to ECG signals for classification purposes. Among these models, CNN is the most common and effective model for arrhythmia detection today \cite{dlreview2020}. However, these methods require an intensive processing pipeline during application and require beat segmentation processes \cite{ji2019electrocardiogram,hwang2020automatic}.

The concept of object detection in computer vision could be applied to arrhythmia detection. Object detection combines image classification and object localisation tasks where a bounding box accompanies the predicted class of an object in real-time \cite{xiao2020review}. In the application of arrhythmia detection, an arrhythmic beat can be treated as an object. Ji et al. \cite{ji2019electrocardiogram} proposed a Faster R-CNN to classify arrhythmia using bounding boxes. However, the Faster R-CNN is a two-stage detector comprising region proposal generation and object detection, thereby increasing architecture complexity and slower detection speeds required for real-time detection. Recently, Hwang et al. \cite{hwang2020automatic} proposed a one-dimensional (1D) CNN You-Only-Look-Once (YOLO)-based model, replacing the bounding box with two bounding windows. Bounding windows with low confidence are discarded during training, losing data in poor-quality samples. Furthermore, post-processing is required to remove multiple bounding windows with the same heartbeat, incurring additional detection time. 

Our study proposes a novel application for arrhythmia detection through the object detection perspective using the state-of-the-art (SOTA) YOLOv8 model. In brief, YOLO is a single-stage detector known for its remarkable object detection speeds with only a forward pass. To our best knowledge, no prior research in object detection literature has focused on detecting arrhythmia using a two-dimensional (2D) YOLO detection model from ECG readings \cite{ji2019electrocardiogram, hwang2020automatic}. Hence, the novel elements of this paper are as follows:
\begin{enumerate}
    \item new object detection-based two-dimensional YOLOv8 arrhythmia detection model;
    \item new methodology in arrhythmia object detection without computationally expensive beat segmentation processes;
    \item preprocessed the one-dimensional MIT-BIH dataset into the two-dimensional images with YOLO annotation format comprising the image, bounding box, and mosaic augmentation; and
    \item improved loss function by introducing dynamic inverse-class frequency to handle class imbalances and implemented Wise IoU to improve precision confidence in poor-quality samples.
\end{enumerate}

We achieved a mAP$_{50}$ of $0.992$ and inference time of $0.002s$ as compared to current SOTA \cite{hwang2020automatic}, which achieved mAP$_{50}$ of $0.960$ with an inference time of $0.03s$ for 5-class classification (shown in Table \ref{mitbih classes}). This model lets users see five classes of detected ECG beats with bounding boxes and prediction confidence. By accurately identifying the individual heartbeats, we can detect any deviation from the regular pattern that can indicate an arrhythmia. Our model could seamlessly integrate into Holter devices with image or video displays for real-time observation and analysis. Moreover, the model and its results can be easily deployed on edge devices and cloud-based systems to facilitate further diagnosis by medical experts. 

\section{Methodology}
\begin{figure}[h!]
    \centering
    \includegraphics[width=\textwidth]{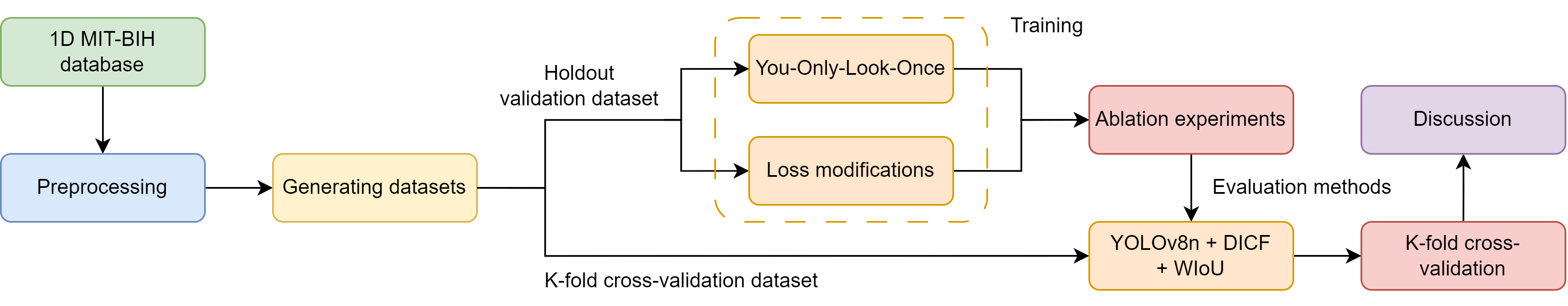}
    \caption{Flow diagram of the proposed methodology} 
    \label{Flow diagram}
\end{figure}

Figure \ref{Flow diagram} illustrates the sequential progression of our paper. The methodology commences with the conversion of the 1D MIT-BIH database into 2D images by plotting waveforms onto a white canvas, followed by annotation to culminate in the final dataset. Our YOLOv8n model was trained on a secluded validation dataset, and we executed ablation experiments employing modifications to both the model and loss functions—comprising inverse-class frequency loss (ICF), dynamic ICF (DICF), and Wise IoU (WIoU). The selection of the most optimal model was based on YOLO evaluation metrics and performance indicators. The model detects each beat on the white canvas and assigns it a class label without needing separate beat segmentation. This selected model was then subjected to a k-fold cross-validation study to ensure the model's generalisability.

In the discussion section, we elucidate the real-time deployment of the model and compare its total detection time with other real-time models. In light of these stages, our research's primary objectives encompass:
\begin{enumerate}
\item Facilitating real-time visual interpretation of ECG waveforms to empower patient self-monitoring and assist physicians in cardiac diagnoses.
\item Proposing a real-time arrhythmia detection model with rapid detection speeds and high classification accuracy.
\item Demonstrating the potential of real-time object detectors within arrhythmia detection applications.
\end{enumerate}

The preprocessing phase (Section \ref{preprocessing}) involved sourcing our dataset from the MIT/Beth Israel Hospital (MIT-BIH) Arrhythmia Database and extracting validated QRS detection points as annotation labels \cite{mitbih2}. We transformed the dataset into full-length patient signals, adhering to the Association for the Advancement of Medical Instrumentation (AAMI) guidelines. This process included the exclusion of four-paced beats and the focal distinction between ventricular ectopic beats (VEBs) and non-ventricular ectopic beats, in line with recommended practices \cite{he2020framework, aami1}. Our study embraced a ground-truth PhysioBank notation scheme.

\subsection{Preprocessing} \label{preprocessing}
We acquired our dataset from the MIT-BIH Arrhythmia Database\footnote{Web page (24.08.2023): \url{https://www.physionet.org/content/mitdb/1.0.0/}} \cite{moody2001impact} and extracted the verified QRS detection points as our ground truth annotation labels \cite{mitbih2}. Subsequently, we transformed the dataset into full-length signals for each patient without resampling and denoising. Following the Association for the Advancement of Medical Instrumentation (AAMI) recommended practice, we removed four-paced beats and focused on distinguishing ventricular ectopic beats (VEBs) from non-ventricular ectopic beats \cite{he2020framework, aami1}. We included a ground-truth PhysioBank annotation symbol above each R-peak in the QRS complex for each patient's full-length signal to annotate the dataset. We extracted $\pm 5$ seconds around non-N waves to rebalance the MIT-BIH dataset, which is heavily skewed towards N-type heartbeats based on the AAMI EC57 standard shown in Table \ref{mitbih classes}. Thus, our preprocessed dataset comprises images with multiple waves of varying numbers, and the model would not be biased towards detecting arrhythmia based on single QRS complexes but instead learn from the patterns in the continuous real-time data.

\begin{table}[ht]
\centering
\caption{MIT-BIH Arrhythmia dataset based on AAMI EC57 standard \cite{aami1}.}
\begin{tabular}{ll}
\toprule[1pt]
\textbf{AAMI Classes} & \textbf{MIT-BIH Arrhythmia Beat Types} \\ \midrule[0.5pt]
\multirow{5}{*}{Normal (N)} & Normal beat (NOR) \\
 & Left bundle branch block (LBBB) \\
 & Right bundle branch block (RBBB) \\
 & Atrial escape beat (AE) \\
 & Nodal escape beat (NE) \\ \midrule[0.25pt]
\multirow{4}{*}{Supraventricular (S)} & Atrial premature beat (AP) \\
 & Aberrant atrial premature beat (aAP) \\
 & Nodal premature beat (NP) \\
 & Supraventricular premature beat (SP) \\ \midrule[0.25pt]
\multirow{2}{*}{Ventricular (V)} & Premature ventricular contraction (PVC) \\
 & Ventricular escape beat (VE) \\ \midrule[0.25pt]
Fusion (F) & Fusion of normal \& ventricular beat (FVN) \\ \midrule[0.25pt]
\multirow{3}{*}{Unknown (Q)} & Paced beat (/) \\
 & Fusion of paced \& normal (FPN) \\
 & Unclassified (U) \\ \bottomrule[1pt]
\end{tabular}
\label{mitbih classes}
\end{table}

\subsection{Generating datasets}
We annotated the preprocessed images with bounding boxes from the base of each R-peak, with each peak being labeled separately with its class label. Each bounding box coordinate is saved in the standard normalised YOLO \textit{xywh} format, where $x$ and $y$ are the box's centre coordinates. Following that, $w$ and $h$ are the width and height of the box encompassing the signal, respectively. Previous research by \cite{ecg_image_data} showed this method effectively extracts RR intervals without filtering or making signal morphology assumptions. We resized our dataset to the YOLO benchmark resolution of 640 by 640 pixels, applied image-level augmentation of 70\% grayscale, and bounding box level augmentations with $\pm 1^{\circ}$ rotation and noise of up to 1\% of pixels \cite{bounding_box_augmentation}. Table \ref{tab:dataset_imbalance} shows the final dataset comprising 42,417 images with an average of 4.343 annotations per image.

\begin{table}[ht]
    \centering
    \caption{Dataset.}
    \begin{tabular}{lll}
    \toprule[1pt]
      \textbf{AAMI Label} & \textbf{Number of Annotations} & \textbf{Composition} \\ \midrule[0.5pt]
      N & 166556 & 45.352\% \\
      V & 88322 & 24.049\% \\
      S & 86861 & 23.652\% \\
      F & 15182 & 4.134\% \\
      Q & 10331 & 2.813\% \\ \midrule[0.5pt]
      Total & 184205 & 100.000\% \\
      \bottomrule[1pt]
    \end{tabular}
    \label{tab:dataset_imbalance}
\end{table}

\subsection{You-Only-Look-Once (YOLO)}

\begin{figure}[h]
    \centering 
    \includegraphics[width=\textwidth]{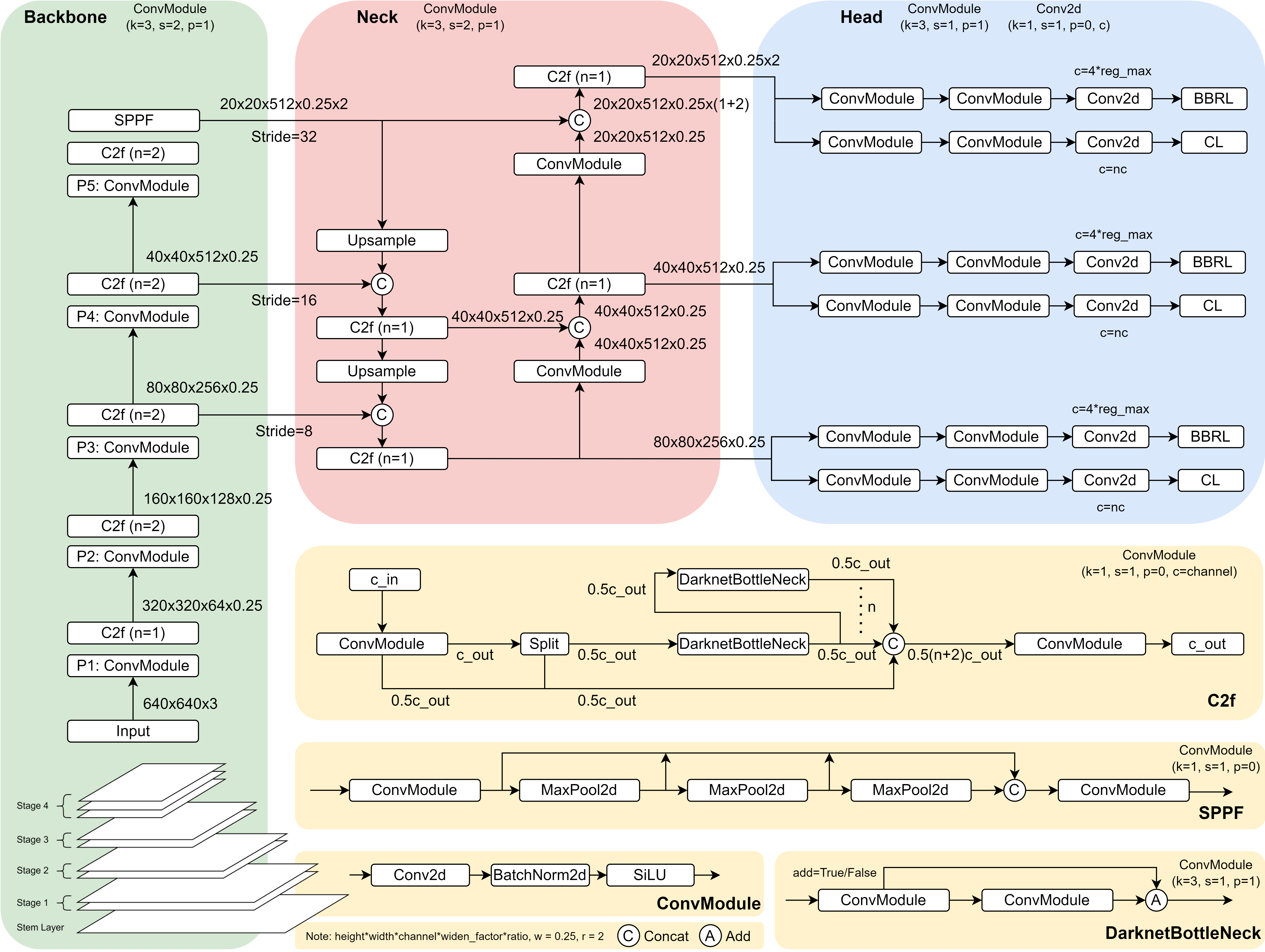}
    \caption{YOLOv8 nano (YOLOv8n) architecture, diagram based on \cite{ terven2023comprehensive, Jocher_YOLO_by_Ultralytics_2023, ju2023fracture}.}
    \label{fig: yolo framework}
\end{figure}

The YOLO framework uses convolutional layers to predict the bounding boxes and the class probabilities of all objects depicted in an image \cite{yolo}. Since the YOLO algorithm is a single-shot detector, it only looks at the image once. YOLO computes a confidence score for each bounding box by multiplying the probability of containing an object in an underlying grid with the Intersection over Union (IoU) between the ground truth and the predicted bounding box. Subsequently, non-maximum suppression (NMS) discards overlapping bounding boxes surrounding the detected object by selecting the box with the highest confidence \cite{terven2023comprehensive,diwan2022object}. In this paper, we used the SOTA YOLOv8 released by Ultralytics in January 2023 \cite{Jocher_YOLO_by_Ultralytics_2023}. The YOLOv8 is an anchor-free model which optimises the number of box predictions, decreasing the time taken for NMS. Figure \ref{fig: yolo framework} shows the overall model architecture of YOLOv8 nano (YOLOv8n), which mainly comprises three parts: Backbone, Neck, and Head. 

\subsubsection{Backbone}
The YOLOv8 model uses a modified CSPDarknet53 backbone, which maintains the idea of Cross Stage Partial (CSP) modules but replaces C3 modules in YOLOv5 with C2f modules to keep the model lightweight for feature extraction. The C2f module combines the C3 module, and ELAN introduced in YOLOv7 \cite{wang2022yolov7}, using the DarknetBottleNeck structure to obtain richer gradient flow information. In Figure \ref{fig: yolo framework}, we denote kernel, stride, and padding as $k$, $s$, and $p$, respectively. $n$ is a parameter of depth that controls the number of BottleNeck stacks by adding more layers, which varies between the model scales: YOLOv8n, YOLOv8s (small), YOLOv8m (medium), YOLOv8l (large), and YOLOv8x (extra large). 

\subsubsection{Neck}
The Backbone connects to the Neck at three different depths (Stage 2, Stage 3, Stage 4) to create a fusion of features obtained from the different layers of the network and passes that to the Head. The Neck comprises path aggregation network (PAN) \cite{pan} and feature pyramid network (FPN) \cite{fpn} structures to prevent information loss due to multiple convolutions. First, the FPN structure upsamples the lower features from the top down, preventing them from losing less location information. In contrast, the PAN structure downsamples the features from the bottom up using ConvModule, allowing the top features to obtain more position information. Like the Backbone, the Neck's FPN structure used the C2f module instead of the C3 module in YOLOv5 and removed ConvModule for direct upsampling. Using the PAN-FPN structure, the output channels of the Neck are equal to the output channels of the Backbone.

\subsubsection{Head} 
The Head comprises three detection Heads, further decoupled into classification and regression tasks. The method of decoupling Heads was first presented in YOLOX and YOLOv6 \cite{terven2023comprehensive} for anchor-free detection. Hence, the model uses task-aligned one-stage object detection (TOOD), $t=s^\alpha \times u^\beta$, where $s$ and $u$ are the classifications and IoU scores, respectively; $\alpha$ and $\beta$ are hyperparameter weights. With task-aligned positive sample matching, $t$ aims to optimise the classification score and IoU for dynamic label assignment by selecting positive samples according to the weighted classification and regression scores. 

\subsubsection{Loss}
\begin{figure}[h!]
    \centering
    \includegraphics{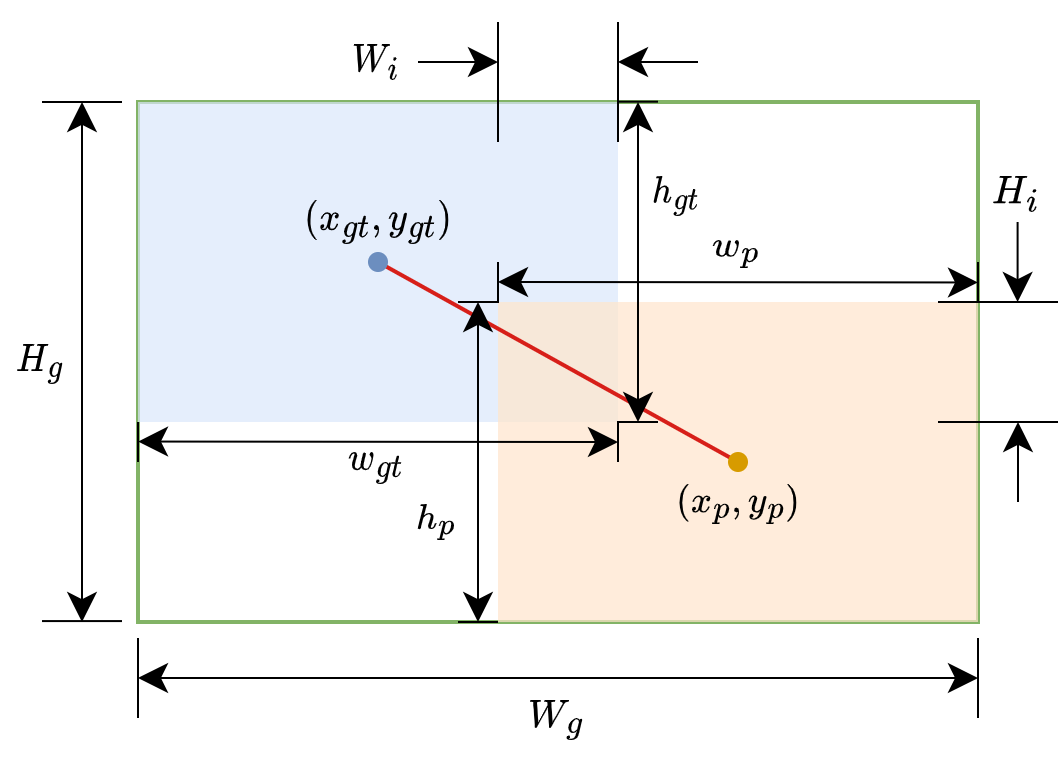}
    \caption{Schematic diagram of bounding box parameters where subscripts $gt$, $p$, $i$, and $g$ denote ground truth, predicted, intersected, and smallest enclosed, respectively. $x$ and $y$ are the coordinates of the bounding boxes.}
    \label{fig: Schematic diagram of bb}
\end{figure}

The loss function is the weighted sum of classification and position-based regression losses. In the original implementation of YOLOv8, specifically the classification loss, the authors replaced  Varifocal Focal Loss (VFL) with the standard Binary Cross Entropy (BCE) loss without explicitly incorporating any weighting scheme defined in Equation (\ref{eq: bce original}) below:
\begin{equation}
    \text{BCE}_\ell = \ell_n(x,y) = y_n\cdot\log(\sigma(x_n)+(1-y_n))\cdot\log(1-\sigma(x_n)),
    \label{eq: bce original}
\end{equation}
where $n$ is the number of samples, $y_n$ is the ground truth value and $x_n$ is the predicted value. The position-based regression losses comprise Distribution Focal Loss (DFL) and Complete IoU (CIoU) loss. Firstly, DFL aims to increase the probability around the target $y$ by rapidly focusing on the target \cite{gfl} and is defined as follows: 
\begin{equation}
    \text{DFL}(S_i, S_{i+1})=-((y_{i+1}-y)\log(S_i)+(y-y_i)\log(S_{i+1})),
    \label{eq: dfl}
\end{equation}
where $y_i$ denotes the left side values of the label $y$, and $y_{i+1}$ denotes the right side values of the label $y$, and $y=\sum^n_{i=0}P(y_i)y_i$ where $P(y_i)$ is implemented using softmax layer denoted by $S_i$. Finally, CIoU loss is defined as the aggregation of IoU, Distance IoU (DIoU) \cite{diou}, and aspect ratio of the prediction and ground truth bounding boxes:
\begin{equation}
    \text{BBRL} = \text{CIoU}_{\ell}=1-\text{IoU}+\frac{d^2}{c^2}+\frac{v^2}{(1-\text{IoU}+v)},
    \label{eq: ciou}
\end{equation}
\begin{equation}
    v=\frac{4}{\pi^2}\left(\arctan\frac{w_{gt}}{h_{gt}}-\arctan\frac{w_p}{h_p}\right)^2,
    \label{eq: v of ciou}
\end{equation}
where $v$ is the measure of the consistency in aspect ratio with $w_{gt}$ and $h_{gt}$ are the width and height of the ground truth bounding box, $w_p$ and $h_p$ are the width and height of the predicted bounding box; $d$ is the Euclidean distance between the centre point of the predicted and ground truth bounding boxes; $c$ is the diagonal length of the smallest box enclosing both boxes \cite{ciou}. Figure \ref{fig: Schematic diagram of bb} shows the bounding box parameters schematic diagram. Finally, the loss function in each decoupled head is expressed as:
\hspace{1em}
\begin{equation}
    f_{\ell}=  \lambda_1 \text{BCE}_\ell + \lambda_2 \text{DFL} + \lambda_3\text{BBRL}.
    \label{eq: total loss}
\end{equation}

\subsection{Loss function modifications}
\subsubsection{Classification}
To address the class imbalance shown in Table \ref{tab:dataset_imbalance}, we introduced dynamic inverse class frequency (DICF) in Equation (\ref{eq: bce multiclass}) and (\ref{eq: bce weights}) to emphasise underrepresented classes. This enhances our model's ability to effectively learn from imbalanced data per batch and improve overall recall (sensitivity) results. 
\begin{equation}
    \text{DICF-BCE}_{\ell}=\ell_{n,c}(x,y) = -w_{n,c}[p_c y_{n,c}\cdot\log(\sigma(x_{n,c})+(1-y_{n,c}))\cdot\log(1-\sigma(x_{n,c}))],
    \label{eq: bce multiclass}
\end{equation}
\begin{equation}
    p_c = \log\left(\frac{n}{f(c)}\right),
    \label{eq: bce weights}
\end{equation}

where $n$ is the total number of samples in the batch, $c$ is the class labels of the annotations in the batch, $f(c)$ denotes the frequency of class $c$, and $p_c$ is the dynamic logarithmic, inverse class frequency weights per batch. In Equation (\ref{eq: bce weights}), $f(c)$ is also updated with respect to mosaic augmentation and albumentations. Finally, we applied a logarithmic transformation to reduce extreme values and prevent any class from dominating the updated weights.
\subsubsection{Bounding box regression}
To improve the precision confidence in the case of low-quality sample data, we modify the BBRL in Equation (\ref{eq: ciou}). We accomplish this by utilising the SOTA Wise-IoU (WIoU) v3 loss to enhance precision for classes with less prominent features or low-quality annotation samples \cite{wiseiou}. WIoU v3 is a two-layer attention-based (WIoU v1) with a dynamic nonmonotonic focusing mechanism (FM) that employs a wise gradient gain allocation approach using an outlier degree $\beta$ of the predicted box. The WIoU v3 can be defined as the WIoU v1 with a nonmonotonic focal number expressed as follows:
\begin{equation}
    \text{BBRL} = \text{WIoU v3}_{\ell}= \gamma \text{WIoU v1}_{\ell},
    \label{eq: wiou v3}
\end{equation}
\begin{equation}
    \gamma= \frac{\beta}{\delta\alpha^{\beta-\delta}},
    \label{eq: gamma and iou}
\end{equation}
\begin{equation}
    \text{WIoU v1}_{\ell}= \text{IoU}_{\ell}\exp{\frac{(x_p-x_{gt})^2+(y_p-y_{gt})^2}{(W^2_g+H^2_g)*}},
    \label{eq: wiou v1}
\end{equation}
\begin{equation}
    \text{IoU}_{\ell} = 1- \text{IoU} = 1 - \frac{W_iH_i}{w_ph_p+w_{gt}h_{gt}-W_iH_i},
    \label{eq: iou}
\end{equation}
where $\alpha$ and $\delta$ are hyperparameters with 1.9 and 3, respectively; $\beta$ is the outlier degree;  and ($*$) denotes detaching from computational graph. During the early stages of training, to prevent low-quality anchor boxes from being dropped, we set a small momentum given by $m = 1 - \sqrt[tn]{0.05}$, where $t$ is the epoch when the lifting speed of AP slows significantly, and $n$ is the number of batches. During later stages of training, when $\beta$ is large, a small gradient gain is assigned to the low-quality anchor boxes to minimise harmful gradients. Hence, WIoU v3 allows the model to mask the influence of low-quality samples while focusing on normal-quality anchor boxes. 

\subsection{Training} \label{training}
\begin{figure}[h!]
    \centering
    \includegraphics[width=0.40\linewidth]{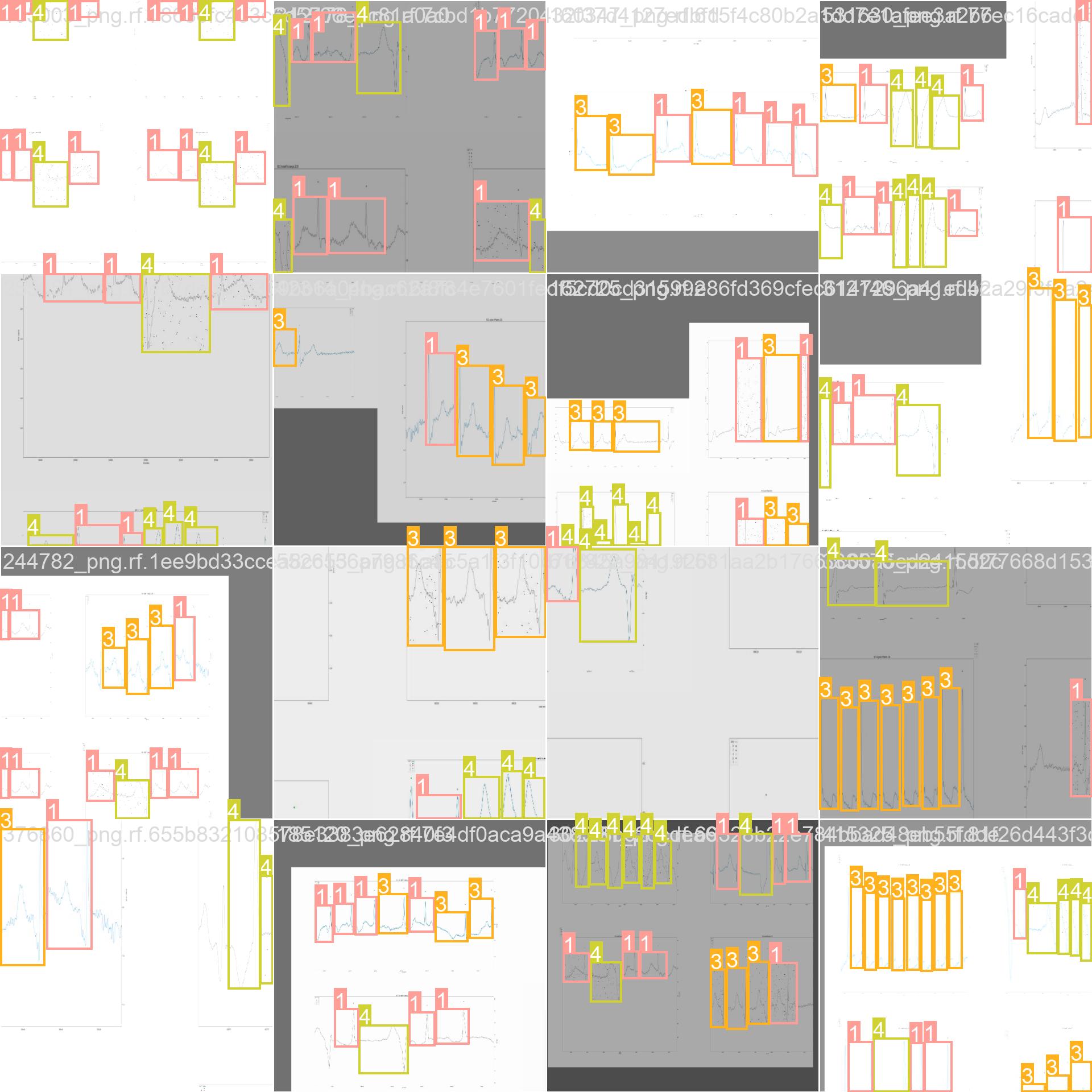} \hspace{1em}%
    \includegraphics[width=0.40\linewidth]{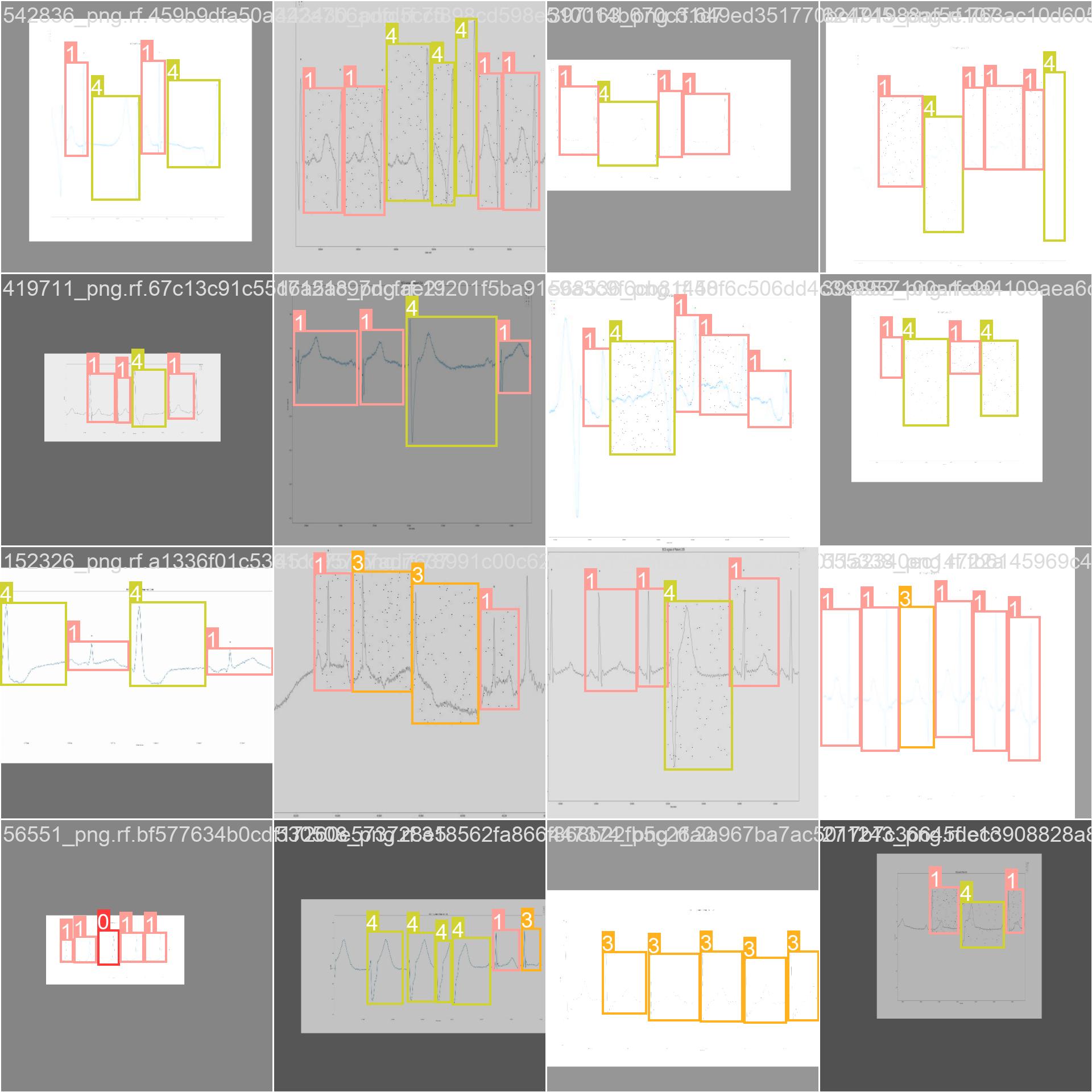}
    \caption{Mosaic augmentation for epochs 1-50 (left) and without mosaic augmentation for epochs 51-200 (right).}
    \label{fig: training dataset}
\end{figure}

In this study, we trained a YOLOv8 nano (YOLOv8n) model without using pretrained weights. The YOLOv8n model comprises only 3.2M parameters with 8.7B FLOPS. We chose this model for its prediction speeds for home-based real-time arrhythmia detection \cite{Jocher_YOLO_by_Ultralytics_2023}. Our model was trained on Tesla V100-SXM2-16GB. The model training parameters are as follows: (1) epochs: 200, (2) optimiser: SGD, (3) batch size: 64, (4) initial learning rate: 0.01, (5) momentum: 0.937, (6) weight decay: 0.0005, (7) IoU threshold: 0.7, (8) warmup epochs: 3.0, (9) warmup momentum: 0.8, (10) NMS: False, (11) cls gain: 2, (12) patience: 50, (13) WIoU$_t$: 15, (14) WIoU$_n$: 515. In addition, we applied mosaic augmentation during the first 50 epochs in training, as shown in Figure \ref{fig: training dataset}. We kept the class balance the same between the train, validation, and test set, with 70\% of the dataset used for training, 20\% for validation, and 10\% for testing for holdout training. 

The model detects each beat within the 10-second window in the image and assigns it a class out of the 5 AAMI classes. This is then validated against the ground truth annotations for training and measuring key metrics. 

\subsection{Evaluation methods}
\subsubsection{YOLO evaluation metrics} \label{yolo metrics}
The performance of YOLO models was evaluated using the mean average precision (mAP) metrics. The mAP is commonly used to measure object detection performance, considering classification and localisation. In addition, the mAP also comprises the precision-recall (PR) area under the curve (AUC), multiple object categories (MOC), and IoU. To balance the PR trade-off, AUC is considered in calculating mAP. For each category, the PR curve is calculated by varying the confidence of the model's prediction. Each class's average precision (AP) is calculated individually by sampling the PR curve to classify and localise multiple ECG categories. Subsequently, the AP at different IoU thresholds is calculated (AP$_{50-90}$), and the mean of APs (mAP$_{50-90}$) for each IoU threshold is calculated across every class \cite{terven2023comprehensive, diwan2022object}. Finally, the overall AP is computed by averaging the AP values calculated at each IoU threshold defined below:
\begin{equation}
\textit{mAP}= \frac{1}{N}\sum_{i = 1}^{N}\textit{AP}_i,
\label{eq: map}
\end{equation}
where $AP_i$ is the average precision of each class, and $N$ is the total number of classes, including the background as a class type. During training, we set the IoU threshold as 0.7, where IoU $\geq$ 0.7 is considered a true positive (TP), while an IoU < 0.7 is classified as a false positive (FP). When the model fails to detect an object where the ground truth exists, it is considered a false negative (FN). On the other hand, a true negative (TN) refers to the remaining image (background) where an object was not detected. 

\subsubsection{Performance metrics} \label{perf metrics}
To further evaluate our results, we defined our classification based on the four possible states: true positive (TP), true negative (TN), false positive (FP), and false negative (FN). Following this, we reported our findings using accuracy, specificity, precision, and recall (sensitivity) and F$_1$, which are standard metrics in arrhythmia classification models calculated as follows:
\begin{equation}
\textit{Accuracy} = \frac{\textit{TP}+\textit{TN}}{\textit{TP}+\textit{TN}+\textit{FP}+\textit{FN}},
\label{eq: accuracy}
\end{equation}

\begin{equation}
\textit{Specificity} = \frac{\textit{TN}}{\textit{TN}+\textit{FP}},
\label{eq: specificity}
\end{equation}

\begin{equation}
\textit{Precision} = \frac{\textit{TP}}{\textit{TP}+\textit{FP}},
\label{eq: precision}
\end{equation}

\begin{equation}
\textit{Recall} = \frac{\textit{TP}}{\textit{TP}+\textit{FN}},
\label{eq: recall}
\end{equation}

\begin{equation}
F_1 = 2 \frac{\textit{precision} \cdot \textit{recall}}{\textit{precision}+\textit{recall}}
\label{eq: f1}
\end{equation}
where accuracy represents the ratio of correct ECG beat classification to all beats; specificity quantifies our model's ability to avoid false positives; precision represents the ratio of correct ECG classification out of all ECG beats predicted as the class; recall (sensitivity) quantifies our model's ability to avoid false negatives by calculating the ratio of genuine arrhythmic ECG beats out of all beats that truly belong to that class, and $F_1$ is the traditional F-measure representing the harmonic mean of precision and recall to provide a balanced assessment of our model's performance \cite{ji2019electrocardiogram,ebrahimi2020review}. 

In the results section, we presented our results based on ablation experiments using holdout training and 10-fold cross-validation. Firstly, we presented the ablation experimental model results for YOLOv8n, YOLOv8n + ICF, YOLOv8n + DICF, YOLOv8n + WIoU, and YOLOv8n + DICF + WIoU. ICF denotes the traditional inverse class frequency method where the weights are initialised once during training instead of updating weights per batch (DICF). Next, we select the best model and present the precision-recall and $F_1$-confidence curves. Secondly, we conducted a 10-fold cross-validation test to ensure the legitimacy of the best model's results, which resamples the dataset into ten equal parts for training and validation. We calculated the performance values of each fold and presented their means and standard deviations. In doing so, we could assess the variation in model performance. Finally, we tabulated the results metrics mentioned above in Equations (\ref{eq: map})$-$(\ref{eq: f1}).

\section{Results}
\subsection{Ablation experiments}
Table \ref{tbl: ablation yolo} shows the ablation experimental results using YOLO evaluation metrics for modification comparison and selecting the best model suited for our application.
\begin{table}[h!]
\centering
\caption{Ablation experimental results using YOLO evaluation metrics (\ref{yolo metrics}).}
\begin{tabular}{llllllll}
\toprule[1pt]
Model & Dataset & P & R & F1-Confidence & mAP$_{50}$ & mAP$_{75}$ & mAP$_{50-90}$ \\ \midrule[0.75pt]
\multirow{2}{*}{YOLOv8n} & Val & 0.981 & 0.974 & 0.977@0.455 & 0.991 & 0.978 & \textbf{0.874} \\
 & Test & 0.980  & 0.972  & 0.976@0.447  & 0.991 & 0.978 & \textbf{0.874} \\ \midrule[0.25pt]
 \multirow{2}{*}{YOLOv8n + ICF} & Val & 0.966 & 0.981 & 0.973@0.422 & 0.991 & 0.968 & 0.871  \\
 & Test & 0.957 & 0.973 & 0.965@0.591 & 0.989 & 0.965 & 0.846 \\ \midrule[0.25pt]
\multirow{2}{*}{YOLOv8n + DICF} & Val & 0.971 & 0.981 & 0.976@0.595 & 0.991 & 0.978 & 0.867 \\
 & Test & 0.969 & 0.978 & 0.973@0.606  & 0.991 & 0.978 &  0.867\\ \midrule[0.25pt]
\multirow{2}{*}{YOLOv8n + WIoU} & Val & \textbf{0.983} & 0.975 & 0.979@0.443 & 0.992 & 0.979 & \textbf{0.874}\\
 & Test & 0.982 & 0.972 & 0.977@0.447 & 0.992 & 0.979 & \textbf{0.874}  \\ \midrule[0.25pt]
\multirow{2}{*}{YOLOv8n + DICF + WIoU} & Val & 0.974 & \textbf{0.981} & \textbf{0.977@0.603} & \textbf{0.992} & \textbf{0.979} & 0.868  \\
 & Test & 0.971  & 0.977 & 0.977@0.589  & 0.992 & 0.979 & 0.869  \\ \bottomrule[1pt]
\end{tabular}
\label{tbl: ablation yolo}
\end{table}

We can observe the following: 
\begin{itemize}
    \item When comparing the unmodified YOLOv8n with YOLOv8 + ICF, there is a significant improvement in recall score but poorer precision, resulting in a poorer mAP@50 score. However, this is expected since $p_c>1$ for the minority classes (F and Q) to improve recall as a trade-off for precision; 
    \item There is a 0.518\% increase in precision when using DICF instead of ICF, which suggests that DICF generally improves the model performance since a varying number of object classes exist in each iteration during training. When using ICF, extreme values exist due to the dataset imbalance in Table \ref{tab:dataset_imbalance}; 
    \item When comparing YOLOv8n with YOLOv8 + WIoU, there is an improvement in mAP@50 and mAP@75 scores, which are consistent to \cite{wiseiou}; 
    \item When comparing YOLOv8n + DICF + WIoU with the other models, it achieved the best scores for Recall and F1-Confidence. Moreover, WIoU improved the precision by 0.309\% against YOLOv8n + DICF and improved confidence among the tested models. 
\end{itemize}

\begin{figure}[h!]
    \centering
    \includegraphics[width=0.9\textwidth]{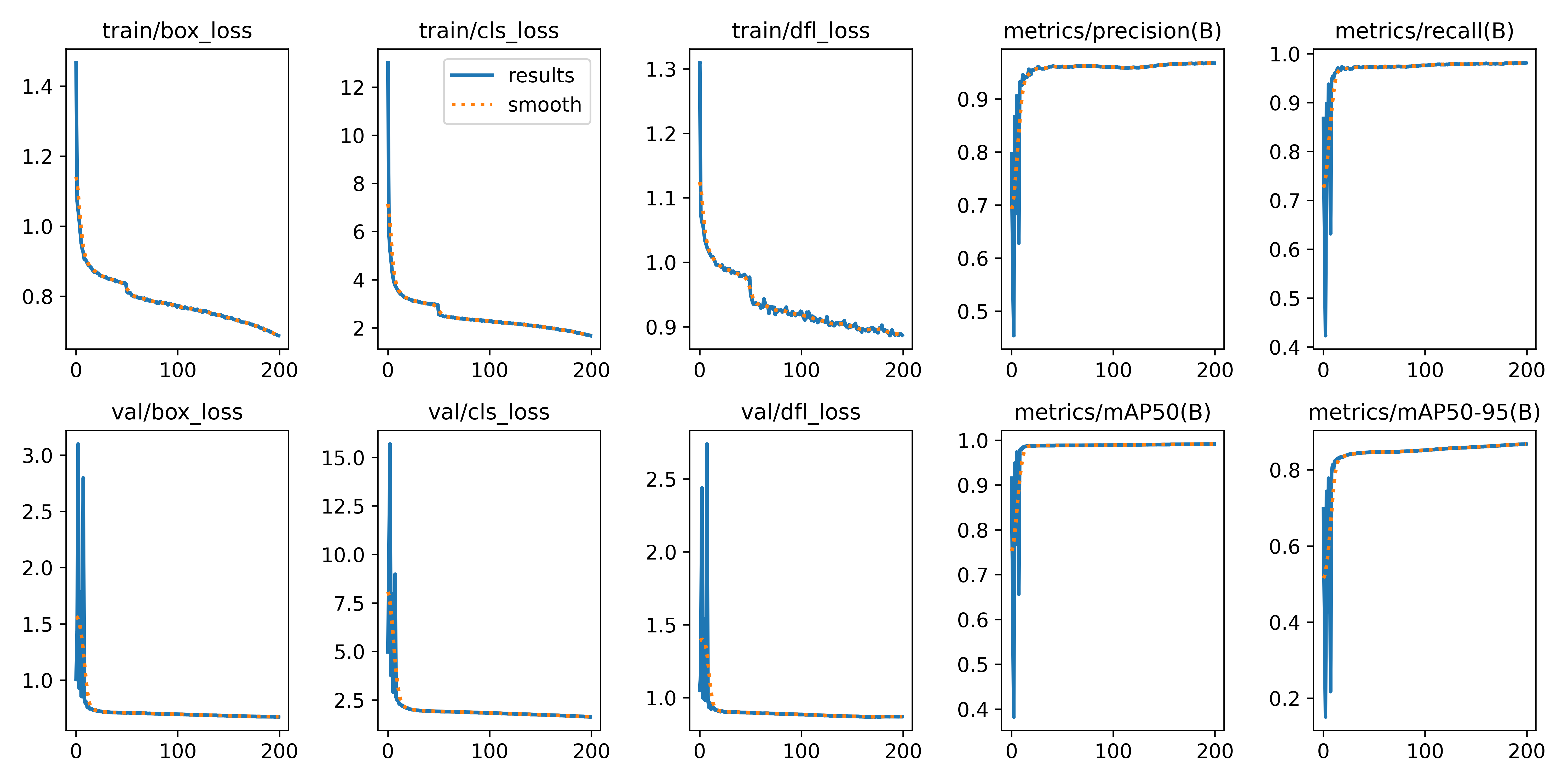}
    \caption{Training and validation results (YOLOv8n + DICF + WIoU).}
    \label{fig:yolo training curve}
\end{figure}

\begin{figure}[h!]
    \centering
    \includegraphics[width=0.45\linewidth]{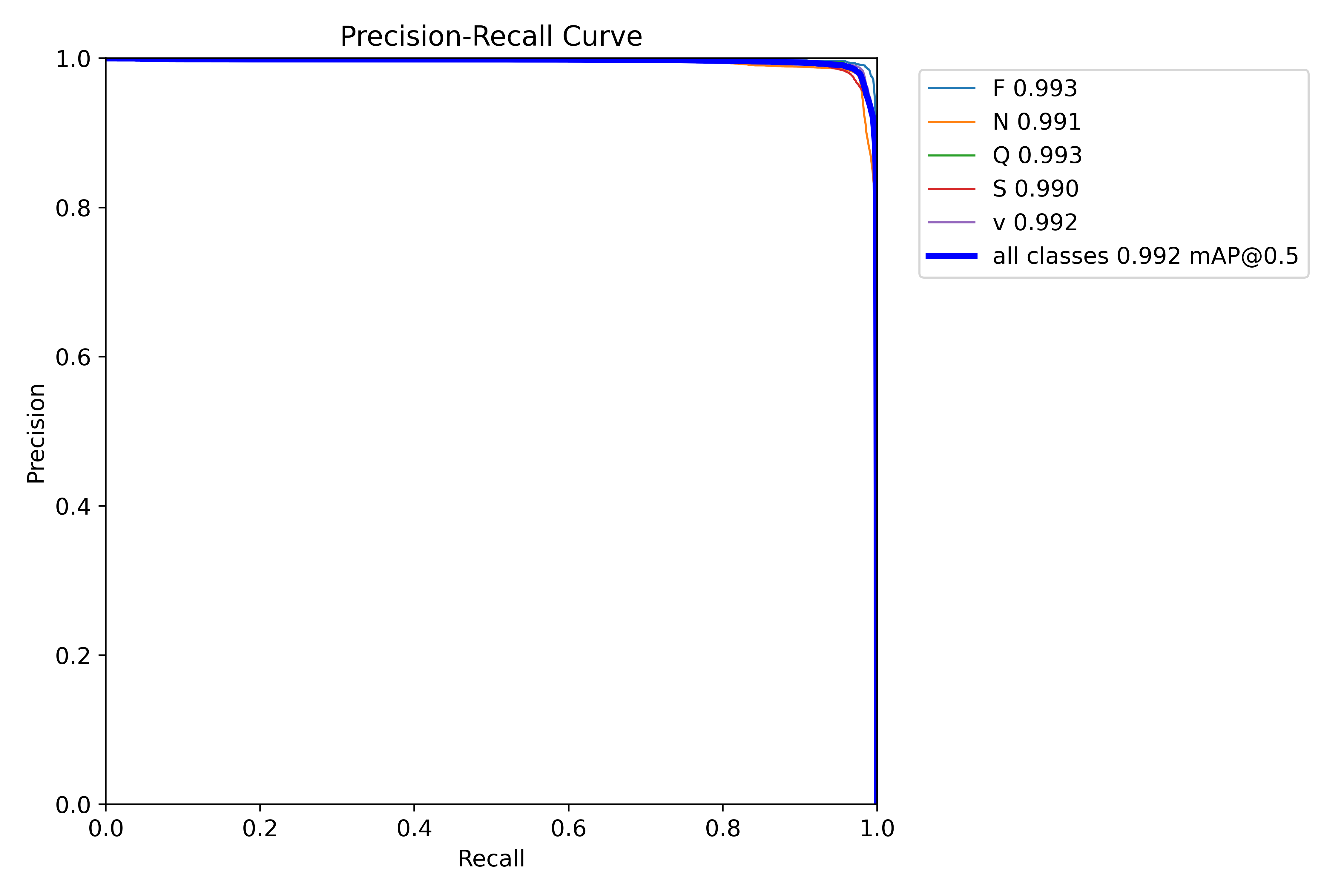}
    \includegraphics[width=0.45\linewidth]{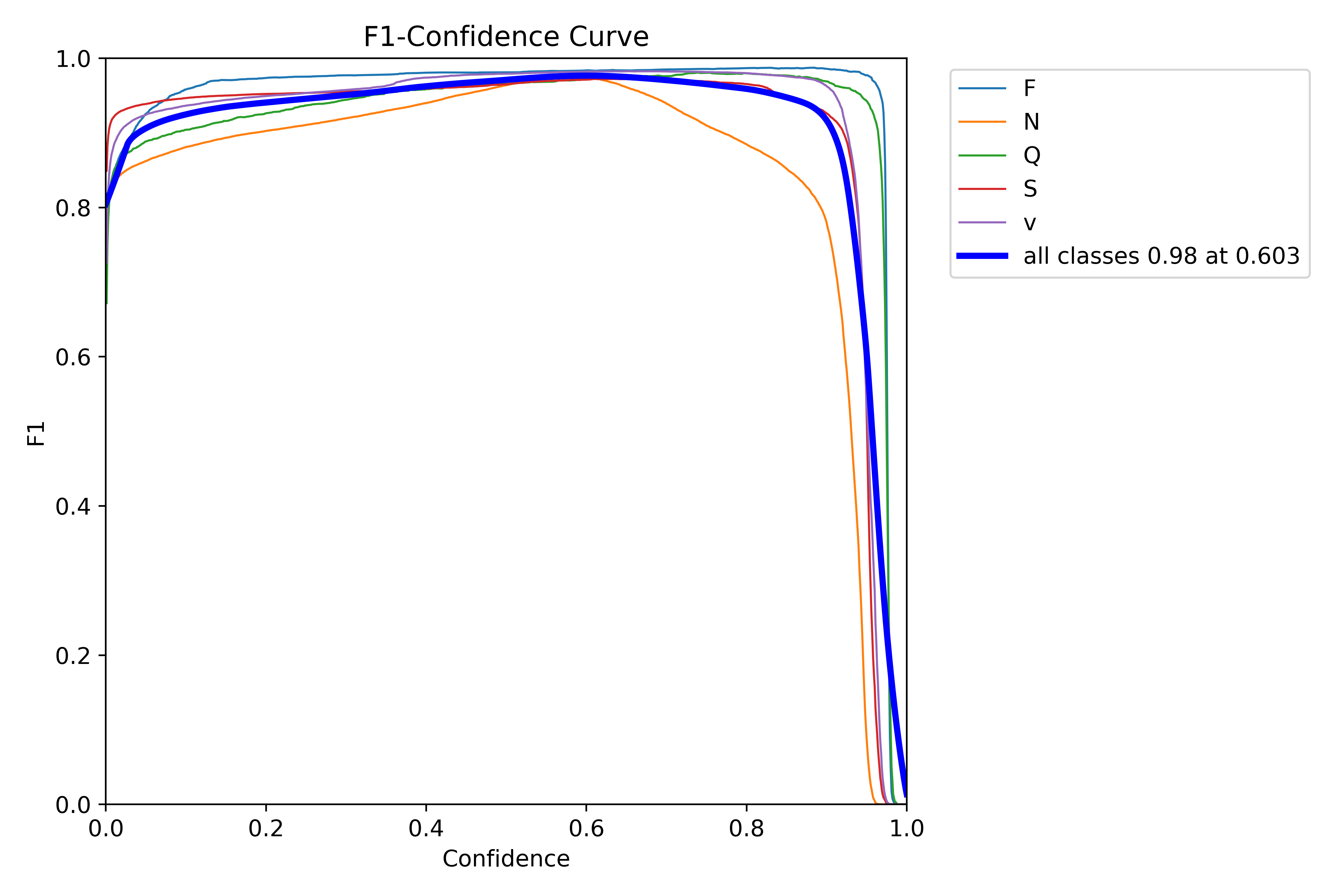}
    \caption{Precision-Recall curve (left) and F1-Confidence curve (right); (YOLOv8n + DICF + WIoU).}
    \label{fig:prf1 curve}
\end{figure}

\begin{figure}[h!]
    \centering
    \includegraphics[width=0.8\linewidth]{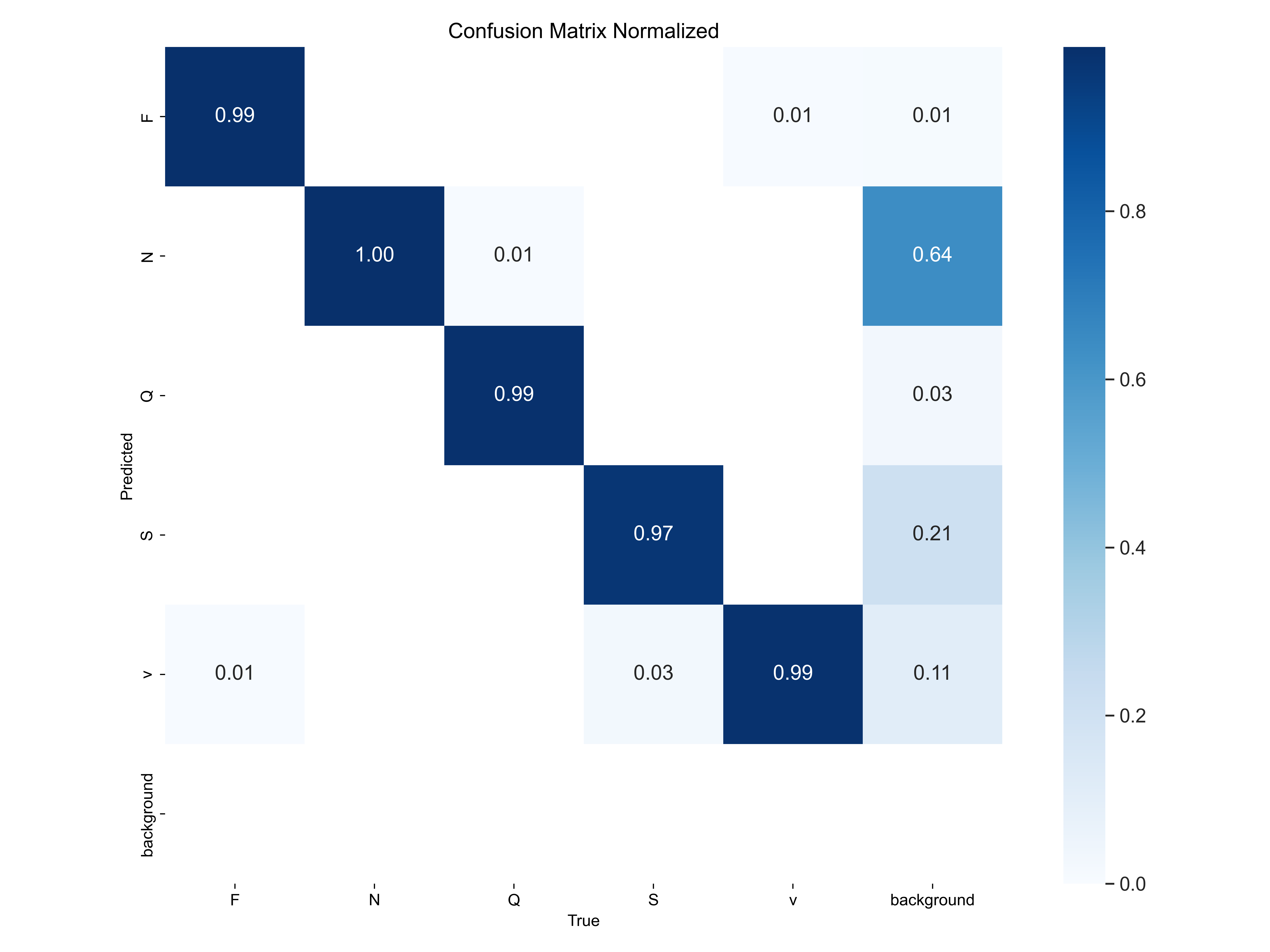}
    \caption{Normalised confusion matrix (YOLOv8n + DICF + WIoU).}
    \label{fig: confusion}
\end{figure}

We selected YOLOv8n + DICF + WIoU from Table \ref{tbl: ablation yolo} for further evaluation as it achieved the best F1-Confidence, mAP@50 and mAP@75. Figure \ref{fig:yolo training curve} shows the model training curve for 200 epochs, where we can see a significant drop across the three losses after closing mosaic augmentations at epoch 51. Figure \ref{fig:prf1 curve} shows the Precision-Recall and F1-Confidence curves, respectively. F and Q achieved the best scores with 0.993 at mAP@0.5 and the best F1 scores with the highest confidence. Figure \ref{fig: confusion} is the normalised confusion matrix. The per-class performance metrics results for the validation and test sets are shown in Table \ref{eval metrics}. The average classification accuracy for validation and test is $99.5\% \pm 0.4\%$ and $99.4\% \pm 0.5\%$, respectively. 

\begin{table}[h!]
\centering
\caption{Performance metrics without background classes (\ref{perf metrics}).}
\begin{tabular}{lllllll}
\toprule[1pt]
\multirow{2}{*}{Classes} & \multirow{2}{*}{Dataset} & \multicolumn{5}{l}{Performance metrics ($\pm$ 0.001)} \\ \cmidrule(lr){3-7} 
 &  & Accuracy & Specificity & Precision & Recall & F1 \\ \midrule[0.5pt]
\multirow{2}{*}{N} & Val & 0.998 & 0.999 & 0.998 & 0.998 & 0.998 \\
 & Test & 0.998 & 0.998 & 0.998 & 0.998 & 0.998 \\ \midrule[0.25pt]
\multirow{2}{*}{V} & Val & 0.989 & 0.989 & 0.966 & 0.988 & 0.977 \\
 & Test & 0.988 & 0.988 & 0.963 & 0.987 & 0.975 \\ \midrule[0.25pt]
\multirow{2}{*}{S} & Val & 0.991 & 0.998 & 0.994 & 0.970 & 0.982 \\
 & Test & 0.990 & 0.998 & 0.993 & 0.965 & 0.978 \\ \midrule[0.25pt]
\multirow{2}{*}{F} & Val & 0.998 & 0.999 &0.975 & 0.984 & 0.979 \\
 & Test & 0.998 & 0.999 & 0.972 & 0.978 & 0.975 \\ \midrule[0.25pt]
\multirow{2}{*}{Q} & Val & 0.999 & 0.999 & 0.990 & 0.995 & 0.992 \\
 & Test & 0.998 & 0.999 & 0.972 & 0.978 & 0.975  \\ \midrule[0.25pt]
\multirow{2}{*}{Average}& Val & \textbf{0.995} & \textbf{0.997} & \textbf{0.985} & \textbf{0.987} & \textbf{0.986} \\
 & Test & \textbf{0.994} & \textbf{0.996} & \textbf{0.980} & \textbf{0.981} & \textbf{0.980} \\
\multirow{2}{*}{Standard deviation} & Val & 0.004 & 0.004 & 0.012 & 0.010 & 0.008 \\ 
 & Test & 0.005 & 0.004 & 0.014 & 0.012 & 0.009 \\ \bottomrule[1pt]
\end{tabular}
\label{eval metrics}
\end{table}

\begin{figure}[h!]
    \centering
    \includegraphics[width=0.8\textwidth]{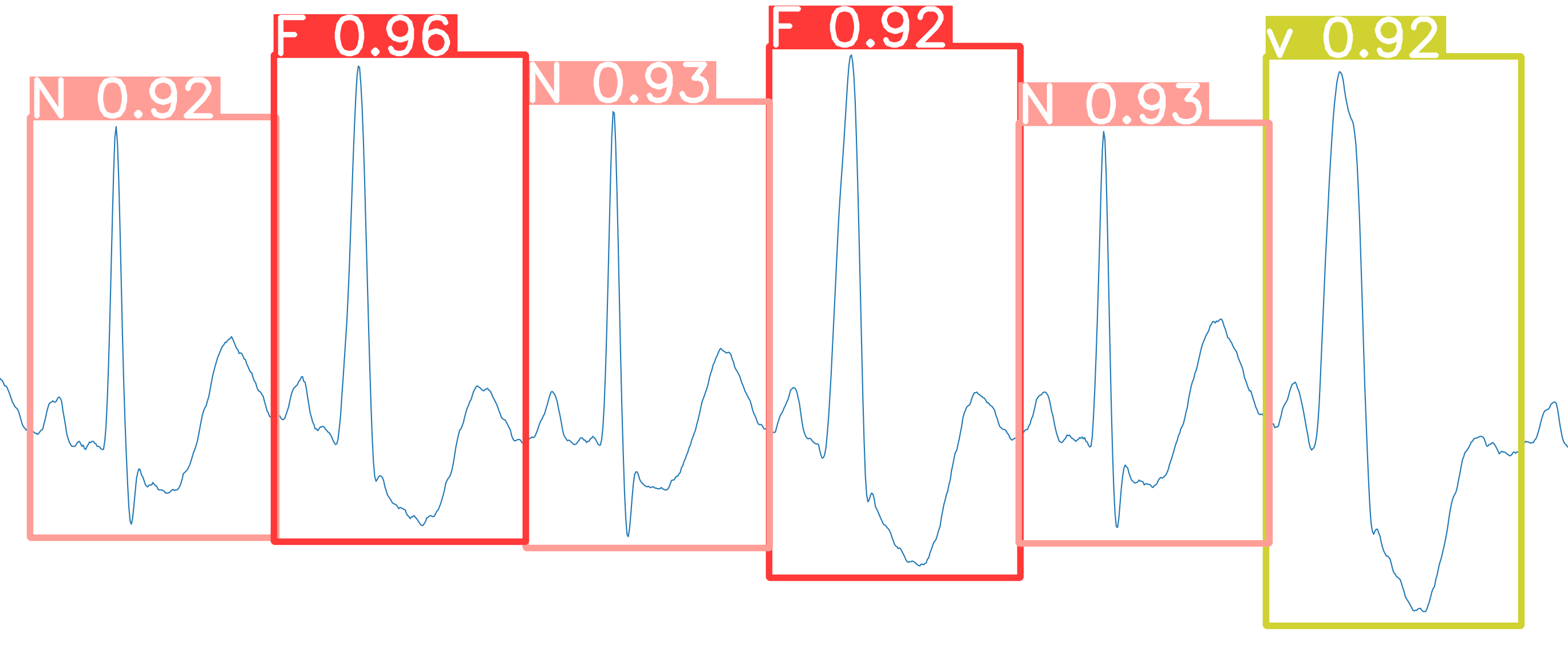}
    \caption{Test set prediction result.}
    \label{offline1}
\end{figure}

Figure \ref{fig:yolo training curve} clearly indicates that the classification and regression losses reached premature convergence when mosaic augmentation was implemented. However, when we closed mosaic augmentation after 50 epochs, the losses continued to improve beyond the 200$^\text{th}$ epoch, and patience was never activated until we stopped training. This suggests our model could be further optimised beyond the 200$^\text{th}$ epoch with WIoU v3. Nonetheless, mosaic augmentation seems appropriate for warming up in the early stages of training. A sample prediction is shown in Figure \ref{offline1}, where our model correctly predicts the beat classes with more than 90\% confidence. Furthermore, we can see that the bounding boxes are precisely drawn without significant overlapping. Therefore, our loss-modified YOLOv8n model can effectively segment between different heartbeats without requiring computationally expensive beat-segmentation processes.

\subsection{K-fold cross-validation} 
To establish the credibility of our training outcomes, we provide 10-fold cross-validation results for YOLOv8n + DICF + WIoU in Tables \ref{10-fold cv yolo eval} and \ref{10-fold cv perf eval}. We observe that the model performs amicably with both YOLO evaluation metrics (\ref{yolo metrics}) and performance metrics (\ref{perf metrics}). Based on YOLO evaluation metrics in Table \ref{10-fold cv yolo eval}, our model achieved averages similar to the holdout training test set, where (a) Precision: $0.969 \pm 0.002$, (b) Recall: $0.979 \pm 0.002$, (c) F1-Confidence: $0.974@0.601 \pm 0.002@0.004$, (d) mAP$_{50}$: $0.992 \pm 0.001$, and (e) mAP$_{50-90}$: $0.870 \pm 0.003$. Each fold has a low standard deviation, indicating consistent model performance.

In relation to classification performance metrics, the average results obtained from the 10-fold cross-validation align consistently with the validation outcomes presented in Table \ref{eval metrics}. Our model achieved an average of (a) $99.5\% \pm 0.1\%$ accuracy, (b) $99.7\% \pm 0.1\%$ specificity, (c) $98.5\% \pm 0.3\%$ precision, (d) $98.7\% \pm 0.2\%$ recall, and (e) $98.6\% \pm 0.2\%$ F1 score. Given the low standard deviations in both results, our model has a generalised understanding of different ECG waveforms and could adapt to new data. 

\begin{table}[h!]
\centering
\caption{Results of 10-fold cross-validation ($\pm$ 0.001) using YOLO evaluation metrics (\ref{yolo metrics}).}
\begin{tabular}{llllll}
\toprule[1pt]
Fold & P & R & F1-Confidence & mAP$_{50}$ & mAP$_{50-90}$ \\ \midrule[0.5pt]
Fold 1 & 0.970 & 0.983 & 0.976@0.604 & 0.992 & 0.868\\
Fold 2 & 0.971 & 0.981 & 0.976@0.602 & 0.992 & 0.868 \\
Fold 3 & 0.971 & 0.978 & 0.974@0.595 & 0.992 & 0.866 \\
Fold 4 & 0.967 & 0.981 & 0.974@0.606 & 0.992 & 0.869 \\
Fold 5 & 0.971 & 0.978 & 0.974@0.595 & 0.992 & 0.867 \\
Fold 6 & 0.966 & 0.978 & 0.972@0.605 & 0.991 & 0.870 \\
Fold 7 & 0.967 & 0.976 & 0.971@0.600 & 0.991 & 0.869 \\
Fold 8 & 0.973 & 0.978 & 0.975@0.597 & 0.992 & 0.870 \\
Fold 9 & 0.968 & 0.979 & 0.973@0.602 & 0.991 & 0.869 \\
Fold 10 & 0.967 & 0.979 & 0.973@0.606 & 0.992 & 0.879 \\ \midrule[0.5pt]
Average & \textbf{0.969} & \textbf{0.979} & \textbf{0.974@0.601} & \textbf{0.992} & \textbf{0.870} \\
Standard Deviation & 0.002 & 0.002 & 0.002@0.004 & 0.001 & 0.003 \\ \bottomrule[1pt]
\end{tabular}
\label{10-fold cv yolo eval}
\end{table}

\begin{table}[h!]
\centering
\caption{Results of 10-fold cross-validation ($\pm$ 0.001) using performance metrics (\ref{perf metrics}).}
\begin{tabular}{llllll}
\toprule[1pt]
Fold & Accuracy & Specificity & Precision & Recall & F1 \\ \midrule[0.5pt]
Fold 1 & 0.994 & 0.996 & 0.979 & 0.985 & 0.982 \\
Fold 2 & 0.997 & 0.998 & 0.988 & 0.991 & 0.989 \\
Fold 3 & 0.996 & 0.998 & 0.987 & 0.989 & 0.988 \\
Fold 4 & 0.995 & 0.997 & 0.985 & 0.987 & 0.986 \\
Fold 5 & 0.996 & 0.997 & 0.986 & 0.989 & 0.987 \\
Fold 6 & 0.995 & 0.997 & 0.983 & 0.987 & 0.985 \\
Fold 7 & 0.994 & 0.996 & 0.981 & 0.984 & 0.982 \\
Fold 8 & 0.995 & 0.997 & 0.987 & 0.987 & 0.987 \\
Fold 9 & 0.996 & 0.997 & 0.988 & 0.986 & 0.987 \\
Fold 10 & 0.995 & 0.997 & 0.984 & 0.987 & 0.985 \\ \midrule[0.5pt]
Average & \textbf{0.995} & \textbf{0.997} & \textbf{0.985} & \textbf{0.987} & \textbf{0.986} \\
Standard Deviation & 0.001 & 0.001 & 0.003 & 0.002 & 0.002 \\ \bottomrule[1pt]
\end{tabular}
\label{10-fold cv perf eval}
\end{table}

\section{Discussion}
Our model has achieved an exceptional arrhythmia classification accuracy of 99.5\% within the realm of object detection. Furthermore, our model excels particularly in detecting the minority classes Q and F, exhibiting the highest overall F1-Confidence scores. By integrating the dynamic inverse-class frequency loss (DICF), we have effectively enhanced recall and F1-Confidence, while the Wise IoU version 3 (WIoU v3) has elevated precision-confidence and mAP scores.

In forthcoming research endeavours, there exists an avenue to explore several class-balancing techniques. These could involve resampling the dataset to under-represent N beats and over-represent F and Q beats or adopting dynamic updates to the effective number of object classes \cite{phan2020resolving}. Additionally, diverse object detection augmentation techniques hold promise, although this pursuit necessitates careful consideration and may entail domain-specific knowledge.

The confusion matrix depicted in Figure \ref{fig: confusion} highlights a notable number of false predictions within the background class. This indicates instances where our model erroneously identified an arrhythmia class when it was, in fact, the image background. This outcome aligns with expectations, given that each beat is solely annotated if the subsequent R-peak is present — simulating continuous signals during real-time detection. Consequently, the model can accurately identify an object class but incurs a penalty for detecting it prematurely. Furthermore, the dataset generation and annotation process plays a role in this phenomenon. With the majority of the ground truth box occupied by the white background, our model could establish a correlation that contributes to lower YOLO evaluation metric scores.

Table \ref{comparison} presents a comparative analysis of our model alongside other relevant works, employing deep learning detectors or CNN models for classification based on the AAMI convention. Our model achieved SOTA results using an object detection-based method to detect arrhythmia. 
Our model excelled in two areas: it exhibited superior performance in terms of YOLO evaluation metrics (\ref{yolo metrics})—specifically, the mAP$_{50}$ score—and demonstrated rapid detection speeds. Furthermore, when it came to classification-based performance metrics (\ref{perf metrics}), our model continued to maintain its lead. It's worth noting that Ji et al. \cite{ji2019electrocardiogram} and Hwang et al. \cite{hwang2020automatic} did not delve into aspects encompassing the total detection time. Nevertheless, in Table \ref{comparison}, we provide a comprehensive overview of their findings for a holistic comparison. For our model, we define the total detection time per frame to include preprocessing, inference, loss, and post-processing, where the inference took 0.7 ms alone. In summary, our loss-modified arrhythmia detector (YOLOv8n + DICF + WIoU v3) achieved 0.992 mAP@50 with an input frame size of 640 pixels with a speed of 430 FPS on Tesla V100-SXM2-16GB.

\begin{table}[h!]
\centering
\caption{Summary and comparison with related work.}
\begin{tabular}{llllllllll}
\toprule[1pt]
\multirow{2}{*}{Work} &
  \multirow{2}{*}{Model} &
  \multicolumn{7}{l}{YOLO evaluation metrics (\ref{yolo metrics}) \& performance metrics (\ref{perf metrics}) $\pm$ 0.001} \\ \cmidrule(lr){3-9} 
 &  & {Accuracy} & {Specificity} & {Precision} & {Recall} & {F1} & {mAP$_{50}$} & {Time} \\ \midrule[0.5pt]
\textit{Kiranyaz et al.} \cite{kiranyaz2015real} & 1D CNN & 0.964 & 0.995 & 0.688 & 0.651 & 0.669 & -  & - s \\
\textit{Xiao et al.}\cite{zhai2018dual} & 2D CNN & 0.984 & 0.978 & 0.659 & 0.721 & 0.681 & - & - s\\
\textit{Ji et al.} \cite{ji2019electrocardiogram} & Faster RCNN & 0.992 & 0.994 & 0.959 & 0.971 & 0.965 & -  & 0.025 s \\
\textit{Hwang et al.} \cite{hwang2020automatic} & 1D YOLO & -  & - & 0.976 & 0.954  & 0.964 & 0.960 & 0.030 s \\ \midrule[0.25pt]
YOLOv8n (Val) & YOLOv8n & \textbf{0.995} & \textbf{0.997} & \textbf{0.985} & \textbf{0.987} &\textbf{0.986} & \textbf{0.992} & \textbf{0.002 s} \\  
YOLOv8n (Test) & YOLOv8n & \textbf{0.994} & \textbf{0.996} & \textbf{0.980} & \textbf{0.981} & \textbf{0.980} & \textbf{0.992} & \textbf{0.002 s} \\ \bottomrule[1pt]
\end{tabular}
\label{comparison}
\end{table}

\subsection{Real-time detection} \label{rt detection}
\begin{figure}[h!]
    \centering
    \includegraphics[width=0.5\linewidth]{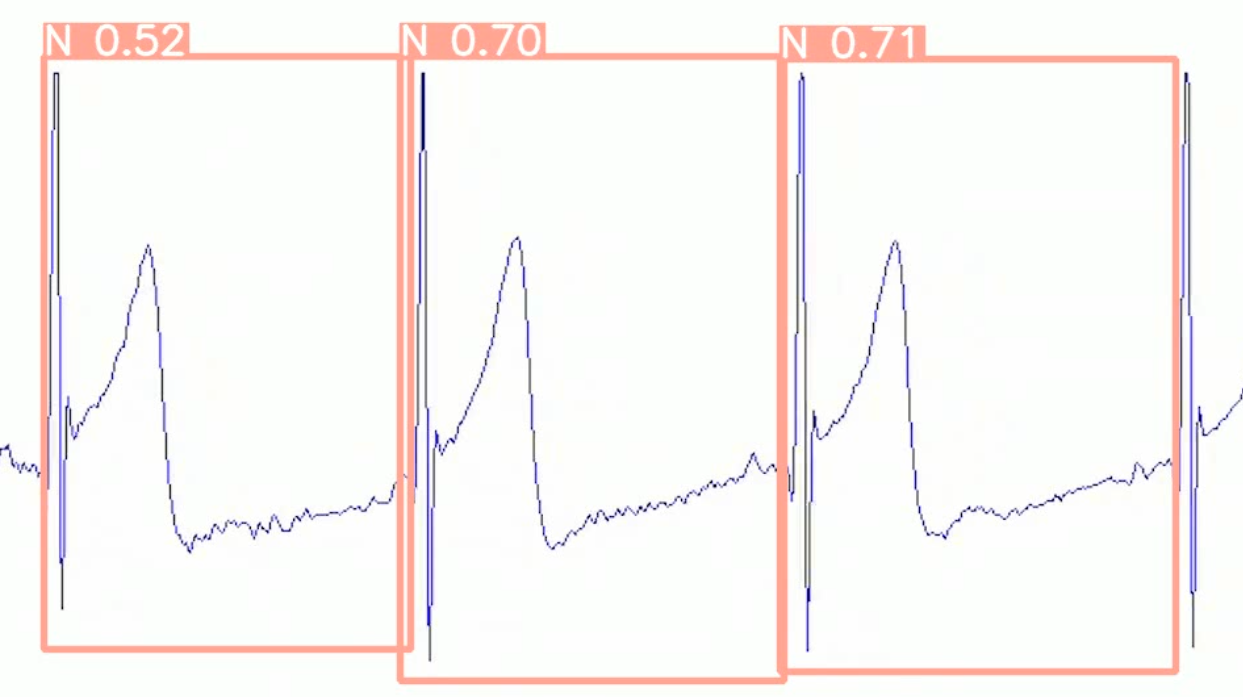}
    \caption{Example of real-time detection.}
    \label{real time example}
\end{figure}

We implemented real-time arrhythmia detection with our trained model using the Sparkfun AD8232. The AD8232 is a commercial cardiograph hardware for biopotential measurement applications in noisy environments. Research from Prasad and Kavanashree \cite{ad8232_ecg} successfully implemented the AD8232 with Internet of Things (IoT) capabilities to monitor the ECG of patients remotely and extended their study to include 12-lead ECG acquisition. We plot the unfiltered serial data from the AD8232 onto a white canvas of 640 by 640 pixels to demonstrate real-time deployability. We fed the updating canvas without digital filters into the trained YOLOv8n model. Figure \ref{real time example} shows a sampled signal captured from the AD8232 through a NodeMCU module as the cost-effective IoT firmware. Here, we used 640 by 640 pixels to reduce preprocessing time.

\subsection{Limitations}
Similar to Ji et al. \cite{ji2019electrocardiogram}, our model also poses the same benefits where we do not require feature extraction and resampling of the original ECG signal. Given the large dataset, we agree with the authors that the model requires long training hours and GPU. However, our work differs in that no denoising and beat segmentation were implemented, simplifying the deep learning pipeline. Notably, using the YOLO family of detectors, our ultimate goal is to balance real-time performance without compromising the detection results. We may improve the results by using larger versions of the YOLOv8, but at a trade-off in real-time performance and incur additional training and deployment costs. This paper used the lightest architecture (YOLOv8n) to demonstrate deployability in Section \ref{rt detection}. Future studies could include using YOLOv8s, YOLOv8m, YOLOv8l, and YOLOv8x, different frame sizes, and exploring the trade-off between speed and accuracy. 

Other limitations may result from the data collection itself, as the portable Holter devices may not be able to provide sufficiently accurate readings. Hence, we expect a decrease in the accuracy upon deployment on edge devices. Moreover, digital filtering for noisy at-home environments may be required, and filtering may result in losing signal features and increased end-to-end processing and inference time. Therefore, further study on preprocessing real-time ECG signals could be conducted to improve detection results. 

While our model achieved SOTA results, high confidence scores, and real-time deployability in object detection-based arrhythmia detection, our study did not consider explainability in its outputs. Interpreting our model's outputs with respect to its input remains challenging due to its architectural complexity and the need for a well-delineated outcome. As a result, the model's decision-making process remains a black box, which can hinder its adoption and acceptance in critical healthcare settings like arrhythmia detection. However, real-time object detection Explainable AI (XAI) tools such as heatmaps, gradient-weighted class activation mapping, saliency maps, and attention mechanisms are in their infancy and do not have native support for arrhythmia detection \cite{loh2022application, xai1, viton2020heatmaps}. Therefore, further research in XAI techniques for object detection-based arrhythmia detection is necessary. 

\subsection{Future works}
This paper could be extended to detect 17 arrhythmia classes \cite{yildirim2018arrhythmia}. We could evaluate the latency of the real-time detection system. Additionally, we could deploy the model in ONNX format using Amazon S3 and Amazon SageMaker endpoints to store data, enhance scalability, and handle inference requests. Hence, we can further apply this research to deploy and develop a fully wireless and portable product at the edge \cite{zhang2023edge, ju2023fracture}. We could also actualise the YOLOv8n model into the cloud system for ambulatory applications, and it is a significant step towards developing real-time applications of XAI models in this domain \cite{loh2022application}. 

Increasing explainability of the model can be explored, as discussed in the sections above. This would allow for increased acceptance in the medical community as the doctors can see why the model is classifying the heartbeat a certain way, and if wrong, could make it easier to tweak or diagnose the errors made by the model.

The model's prediction of arrhythmia class on the background can be improved by varying the background across images or training the model on a transparent background image instead, which could reduce possible memorization occurring in the model, leading to it wrongly classifying white backgrounds as arrhythmia classes. Lastly, we could fine-tune the models with patient data from the portable Holter devices to improve accuracy on data from edge devices.

\section{Conclusion}
In this paper, we implemented a loss-modified YOLOv8 model for patient self-monitoring, providing users with visual feedback on ECG signals with detected classes and their prediction confidence. We trained the YOLOv8n model using the MIT-BIH dataset and improved the results of the minority classes using dynamic inverse class frequency and Wise IoU. Our model achieved state-of-the-art results for arrhythmia detection in the lens of object detection, with an average accuracy of 99.4\% and 99.5\% for validation and test sets, respectively. Moreover, these results are consistent with 10-fold cross-validation tests. Therefore, the YOLOv8 model is suitable for applying real-time arrhythmia detection. While the field is relatively young, we have demonstrated the potential of YOLOv8 as a real-time application for analysing ECG signals. This development could have significant implications for patient self-monitoring, providing valuable insights into heart health and enabling early detection of potentially life-threatening conditions. Future research could focus on explainability and clinical trials to validate its performance in real-world scenarios. 

\section*{Acknowledgements}
We thank Professor U Rajendra Acharya and Associate Professor Oliver Faust for their review of our work. We extend our gratitude to Associate Professor Chua Kian Jon Ernest and Mr Nelliyan Karuppiah for guiding us and providing valuable advice in developing the real-time detection device presented in Chapter \ref{rt detection}.

\clearpage
\bibliographystyle{unsrt}  
\bibliography{references}  

\begin{thebibliography}{10}

\bibitem{cdc_database}
{Centers for Disease Control and Prevention, National Center for Health
  Statistics}.
\newblock About multiple cause of death, 1999–2019, 2021.
\newblock data retrieved from CDC WONDER Online Database website,
  \url{https://wonder.cdc.gov/mcd-icd10.html}.

\bibitem{hammad2021automated}
Mohamed Hammad, Rajesh~NVPS Kandala, Amira Abdelatey, Moloud Abdar, Mariam
  Zomorodi-Moghadam, Ru~San~Tan, U~Rajendra Acharya, Joanna P{\l}awiak, Ryszard
  Tadeusiewicz, Vladimir Makarenkov, et~al.
\newblock Automated detection of shockable ecg signals: A review.
\newblock {\em Information Sciences}, 571:580--604, 2021.

\bibitem{grabowski2022classification}
Bartosz Grabowski, Przemys{\l}aw G{\l}omb, Wojciech Masarczyk, Pawe{\l}
  P{\l}awiak, {\"O}zal Y{\i}ld{\i}r{\i}m, U~Rajendra Acharya, and Ru-San Tan.
\newblock Classification and self-supervised regression of arrhythmic ecg
  signals using convolutional neural networks.
\newblock {\em arXiv preprint arXiv:2210.14253}, 2022.

\bibitem{DOWD2014}
F.J. Dowd.
\newblock Arrhythmias.
\newblock In {\em Reference Module in Biomedical Sciences}. Elsevier, 2014.

\bibitem{pham2021novel}
The-Hanh Pham, Vinitha Sree, John Mapes, Sumeet Dua, Oh~Shu Lih, Joel~EW Koh,
  Edward~J Ciaccio, and U~Rajendra Acharya.
\newblock A novel machine learning framework for automated detection of
  arrhythmias in ecg segments.
\newblock {\em Journal of Ambient Intelligence and Humanized Computing}, pages
  1--18, 2021.

\bibitem{yildirim2018arrhythmia}
{\"O}zal Y{\i}ld{\i}r{\i}m, Pawe{\l} P{\l}awiak, Ru-San Tan, and U~Rajendra
  Acharya.
\newblock Arrhythmia detection using deep convolutional neural network with
  long duration ecg signals.
\newblock {\em Computers in biology and medicine}, 102:411--420, 2018.

\bibitem{martis2013application}
Roshan~Joy Martis, U~Rajendra Acharya, Choo~Min Lim, KM~Mandana, Ajoy~K Ray,
  and Chandan Chakraborty.
\newblock Application of higher order cumulant features for cardiac health
  diagnosis using ecg signals.
\newblock {\em International journal of neural systems}, 23(04):1350014, 2013.

\bibitem{mathews2018novel}
Sherin~M Mathews, Chandra Kambhamettu, and Kenneth~E Barner.
\newblock A novel application of deep learning for single-lead ecg
  classification.
\newblock {\em Computers in biology and medicine}, 99:53--62, 2018.

\bibitem{liu2022multiclass}
Zengding Liu, Bin Zhou, Zhiming Jiang, Xi~Chen, Ye~Li, Min Tang, and Fen Miao.
\newblock Multiclass arrhythmia detection and classification from
  photoplethysmography signals using a deep convolutional neural network.
\newblock {\em Journal of the American Heart Association}, 11(7):e023555, 2022.

\bibitem{luz2016ecg}
Eduardo Jos{\'e} da~S Luz, William~Robson Schwartz, Guillermo
  C{\'a}mara-Ch{\'a}vez, and David Menotti.
\newblock Ecg-based heartbeat classification for arrhythmia detection: A
  survey.
\newblock {\em Computer methods and programs in biomedicine}, 127:144--164,
  2016.

\bibitem{feather2020kumar}
Adam Feather, David Randall, and Mona Waterhouse.
\newblock {\em Kumar and clark's clinical medicine E-Book}.
\newblock Elsevier Health Sciences, 2020.

\bibitem{lilly2012pathophysiology}
Leonard~S Lilly.
\newblock {\em Pathophysiology of heart disease: a collaborative project of
  medical students and faculty}.
\newblock Lippincott Williams \& Wilkins, 2012.

\bibitem{bigler2021accuracy}
Marius~Reto Bigler, Patrick Zimmermann, Athanasios Papadis, and Christian
  Seiler.
\newblock Accuracy of intracoronary ecg parameters for myocardial ischemia
  detection.
\newblock {\em Journal of electrocardiology}, 64:50--57, 2021.

\bibitem{prabhakararao2020myocardial}
Eedara Prabhakararao and Samarendra Dandapat.
\newblock Myocardial infarction severity stages classification from ecg signals
  using attentional recurrent neural network.
\newblock {\em IEEE Sensors Journal}, 20(15):8711--8720, 2020.

\bibitem{lyakhov2021system}
Pavel Lyakhov, Mariya Kiladze, and Ulyana Lyakhova.
\newblock System for neural network determination of atrial fibrillation on ecg
  signals with wavelet-based preprocessing.
\newblock {\em Applied Sciences}, 11(16):7213, 2021.

\bibitem{glass2006cardiac}
Leon Glass.
\newblock Cardiac oscillations and arrhythmia analysis.
\newblock {\em Complex Systems Science in Biomedicine}, pages 409--422, 2006.

\bibitem{loh2022application}
Hui~Wen Loh, Chui~Ping Ooi, Silvia Seoni, Prabal~Datta Barua, Filippo Molinari,
  and U~Rajendra Acharya.
\newblock Application of explainable artificial intelligence for healthcare: A
  systematic review of the last decade (2011--2022).
\newblock {\em Computer Methods and Programs in Biomedicine}, page 107161,
  2022.

\bibitem{hermitefunc}
Martin Lagerholm, Carsten Peterson, Guido Braccini, Lars Edenbrandt, and Leif
  Sornmo.
\newblock Clustering ecg complexes using hermite functions and self-organizing
  maps.
\newblock {\em IEEE Transactions on Biomedical Engineering}, 47(7):838--848,
  2000.

\bibitem{wavelettransform}
Lotfi Senhadji, G~Carrault, JJ~Bellanger, and Gianfranco Passariello.
\newblock Comparing wavelet transforms for recognizing cardiac patterns.
\newblock {\em IEEE Engineering in Medicine and Biology Magazine},
  14(2):167--173, 1995.

\bibitem{hu1997patient}
Yu~Hen Hu, Surekha Palreddy, and Willis~J Tompkins.
\newblock A patient-adaptable ecg beat classifier using a mixture of experts
  approach.
\newblock {\em IEEE transactions on biomedical engineering}, 44(9):891--900,
  1997.

\bibitem{svm1}
J~Millet-Roig, R~Ventura-Galiano, FJ~Chorro-Gasco, and A~Cebrian.
\newblock Support vector machine for arrhythmia discrimination with wavelet
  transform-based feature selection.
\newblock In {\em Computers in Cardiology 2000. Vol. 27 (Cat. 00CH37163)},
  pages 407--410. IEEE, 2000.

\bibitem{knn1}
Yasin Kaya and H{\"u}seyin Pehlivan.
\newblock Classification of premature ventricular contraction in ecg.
\newblock {\em International Journal of Advanced Computer Science and
  Applications}, 6(7), 2015.

\bibitem{knn2}
I~Christov, I~Jekova, and G~Bortolan.
\newblock Premature ventricular contraction classification by the kth
  nearest-neighbours rule.
\newblock {\em Physiological measurement}, 26(1):123, 2005.

\bibitem{decision3}
Santanu Sahoo, Asit Subudhi, Manasa Dash, and Sukanta Sabut.
\newblock Automatic classification of cardiac arrhythmias based on hybrid
  features and decision tree algorithm.
\newblock {\em International Journal of Automation and Computing},
  17(4):551--561, 2020.

\bibitem{lecun}
Yann LeCun, Yoshua Bengio, and Geoffrey Hinton.
\newblock Deep learning.
\newblock {\em nature}, 521(7553):436--444, 2015.

\bibitem{oh2018automated}
Shu~Lih Oh, Eddie~YK Ng, Ru~San~Tan, and U~Rajendra Acharya.
\newblock Automated diagnosis of arrhythmia using combination of cnn and lstm
  techniques with variable length heart beats.
\newblock {\em Computers in biology and medicine}, 102:278--287, 2018.

\bibitem{acharya2017deep}
U~Rajendra Acharya, Shu~Lih Oh, Yuki Hagiwara, Jen~Hong Tan, Muhammad Adam,
  Arkadiusz Gertych, and Ru~San~Tan.
\newblock A deep convolutional neural network model to classify heartbeats.
\newblock {\em Computers in biology and medicine}, 89:389--396, 2017.

\bibitem{dlreview2020}
Zahra Ebrahimi, Mohammad Loni, Masoud Daneshtalab, and Arash Gharehbaghi.
\newblock A review on deep learning methods for ecg arrhythmia classification.
\newblock {\em Expert Systems with Applications: X}, 7:100033, 2020.

\bibitem{ji2019electrocardiogram}
Yinsheng Ji, Sen Zhang, and Wendong Xiao.
\newblock Electrocardiogram classification based on faster regions with
  convolutional neural network.
\newblock {\em Sensors}, 19(11):2558, 2019.

\bibitem{hwang2020automatic}
Won~Hee Hwang, Chan~Hee Jeong, Dong~Hyun Hwang, and Young~Chang Jo.
\newblock Automatic detection of arrhythmias using a yolo-based network with
  long-duration ecg signals.
\newblock {\em Engineering Proceedings}, 2(1):84, 2020.

\bibitem{xiao2020review}
Youzi Xiao, Zhiqiang Tian, Jiachen Yu, Yinshu Zhang, Shuai Liu, Shaoyi Du, and
  Xuguang Lan.
\newblock A review of object detection based on deep learning.
\newblock {\em Multimedia Tools and Applications}, 79:23729--23791, 2020.

\bibitem{mitbih2}
George~B Moody, Roger~G Mark, and Ary~L Goldberger.
\newblock Physionet: a web-based resource for the study of physiologic signals.
\newblock {\em IEEE Engineering in Medicine and Biology Magazine},
  20(3):70--75, 2001.

\bibitem{he2020framework}
Jinyuan He, Jia Rong, Le~Sun, Hua Wang, Yanchun Zhang, and Jiangang Ma.
\newblock A framework for cardiac arrhythmia detection from iot-based ecgs.
\newblock {\em World Wide Web}, 23:2835--2850, 2020.

\bibitem{aami1}
AAMI ECAR.
\newblock Recommended practice for testing and reporting performance results of
  ventricular arrhythmia detection algorithms.
\newblock {\em Association for the Advancement of Medical Instrumentation}, 69,
  1987.

\bibitem{moody2001impact}
George~B Moody and Roger~G Mark.
\newblock The impact of the mit-bih arrhythmia database.
\newblock {\em IEEE engineering in medicine and biology magazine},
  20(3):45--50, 2001.

\bibitem{ecg_image_data}
Mohammad Kachuee, Shayan Fazeli, and Majid Sarrafzadeh.
\newblock Ecg heartbeat classification: A deep transferable representation.
\newblock In {\em 2018 IEEE international conference on healthcare informatics
  (ICHI)}, pages 443--444. IEEE, 2018.

\bibitem{bounding_box_augmentation}
Barret Zoph, Ekin~D Cubuk, Golnaz Ghiasi, Tsung-Yi Lin, Jonathon Shlens, and
  Quoc~V Le.
\newblock Learning data augmentation strategies for object detection.
\newblock In {\em Computer Vision--ECCV 2020: 16th European Conference,
  Glasgow, UK, August 23--28, 2020, Proceedings, Part XXVII 16}, pages
  566--583. Springer, 2020.

\bibitem{yolo}
Joseph Redmon, Santosh Divvala, Ross Girshick, and Ali Farhadi.
\newblock You only look once: Unified, real-time object detection.
\newblock In {\em Proceedings of the IEEE conference on computer vision and
  pattern recognition}, pages 779--788, 2016.

\bibitem{terven2023comprehensive}
Juan Terven and Diana Cordova-Esparza.
\newblock A comprehensive review of yolo: From yolov1 to yolov8 and beyond.
\newblock {\em arXiv preprint arXiv:2304.00501}, 2023.

\bibitem{diwan2022object}
Tausif Diwan, G~Anirudh, and Jitendra~V Tembhurne.
\newblock Object detection using yolo: Challenges, architectural successors,
  datasets and applications.
\newblock {\em Multimedia Tools and Applications}, pages 1--33, 2022.

\bibitem{Jocher_YOLO_by_Ultralytics_2023}
Glenn Jocher, Ayush Chaurasia, and Jing Qiu.
\newblock {YOLO by Ultralytics}, January 2023.

\bibitem{ju2023fracture}
Rui-Yang Ju and Weiming Cai.
\newblock Fracture detection in pediatric wrist trauma x-ray images using
  yolov8 algorithm.
\newblock {\em arXiv preprint arXiv:2304.05071}, 2023.

\bibitem{wang2022yolov7}
Chien-Yao Wang, Alexey Bochkovskiy, and Hong-Yuan~Mark Liao.
\newblock {YOLOv7}: Trainable bag-of-freebies sets new state-of-the-art for
  real-time object detectors.
\newblock {\em arXiv preprint arXiv:2207.02696}, 2022.

\bibitem{pan}
Shu Liu, Lu~Qi, Haifang Qin, Jianping Shi, and Jiaya Jia.
\newblock Path aggregation network for instance segmentation.
\newblock In {\em Proceedings of the IEEE conference on computer vision and
  pattern recognition}, pages 8759--8768, 2018.

\bibitem{fpn}
Tsung-Yi Lin, Piotr Doll{\'a}r, Ross Girshick, Kaiming He, Bharath Hariharan,
  and Serge Belongie.
\newblock Feature pyramid networks for object detection.
\newblock In {\em Proceedings of the IEEE conference on computer vision and
  pattern recognition}, pages 2117--2125, 2017.

\bibitem{gfl}
Xiang Li, Wenhai Wang, Lijun Wu, Shuo Chen, Xiaolin Hu, Jun Li, Jinhui Tang,
  and Jian Yang.
\newblock Generalized focal loss: Learning qualified and distributed bounding
  boxes for dense object detection.
\newblock {\em Advances in Neural Information Processing Systems},
  33:21002--21012, 2020.

\bibitem{diou}
Zhaohui Zheng, Ping Wang, Wei Liu, Jinze Li, Rongguang Ye, and Dongwei Ren.
\newblock Distance-iou loss: Faster and better learning for bounding box
  regression.
\newblock In {\em Proceedings of the AAAI conference on artificial
  intelligence}, volume~34, pages 12993--13000, 2020.

\bibitem{ciou}
Zhaohui Zheng, Ping Wang, Dongwei Ren, Wei Liu, Rongguang Ye, Qinghua Hu, and
  Wangmeng Zuo.
\newblock Enhancing geometric factors in model learning and inference for
  object detection and instance segmentation.
\newblock {\em IEEE transactions on cybernetics}, 52(8):8574--8586, 2021.

\bibitem{wiseiou}
Zanjia Tong, Yuhang Chen, Zewei Xu, and Rong Yu.
\newblock Wise-iou: Bounding box regression loss with dynamic focusing
  mechanism.
\newblock {\em arXiv preprint arXiv:2301.10051}, 2023.

\bibitem{ebrahimi2020review}
Zahra Ebrahimi, Mohammad Loni, Masoud Daneshtalab, and Arash Gharehbaghi.
\newblock A review on deep learning methods for ecg arrhythmia classification.
\newblock {\em Expert Systems with Applications: X}, 7:100033, 2020.

\bibitem{phan2020resolving}
Trong~Huy Phan and Kazuma Yamamoto.
\newblock Resolving class imbalance in object detection with weighted cross
  entropy losses.
\newblock {\em arXiv preprint arXiv:2006.01413}, 2020.

\bibitem{kiranyaz2015real}
Serkan Kiranyaz, Turker Ince, and Moncef Gabbouj.
\newblock Real-time patient-specific ecg classification by 1-d convolutional
  neural networks.
\newblock {\em IEEE Transactions on Biomedical Engineering}, 63(3):664--675,
  2015.

\bibitem{zhai2018dual}
Xiaolong Zhai and Chung Tin.
\newblock Automated ecg classification using dual heartbeat coupling based on
  convolutional neural network.
\newblock {\em IEEE Access}, 6:27465--27472, 2018.

\bibitem{ad8232_ecg}
Anitha~S Prasad and N~Kavanashree.
\newblock Ecg monitoring system using ad8232 sensor.
\newblock In {\em 2019 International Conference on Communication and
  Electronics Systems (ICCES)}, pages 976--980. IEEE, 2019.

\bibitem{xai1}
Alejandro~Barredo Arrieta, Natalia D{\'\i}az-Rodr{\'\i}guez, Javier Del~Ser,
  Adrien Bennetot, Siham Tabik, Alberto Barbado, Salvador Garc{\'\i}a, Sergio
  Gil-L{\'o}pez, Daniel Molina, Richard Benjamins, et~al.
\newblock Explainable artificial intelligence (xai): Concepts, taxonomies,
  opportunities and challenges toward responsible ai.
\newblock {\em Information fusion}, 58:82--115, 2020.

\bibitem{viton2020heatmaps}
Fabien Viton, Mahmoud Elbattah, Jean-Luc Gu{\'e}rin, and Gilles Dequen.
\newblock Heatmaps for visual explainability of cnn-based predictions for
  multivariate time series with application to healthcare.
\newblock In {\em 2020 IEEE International Conference on Healthcare Informatics
  (ICHI)}, pages 1--8. IEEE, 2020.

\bibitem{zhang2023edge}
Shihao Zhang, Hekai Yang, Chunhua Yang, Wenxia Yuan, Xinghui Li, Xinghua Wang,
  Yinsong Zhang, Xiaobo Cai, Yubo Sheng, Xiujuan Deng, et~al.
\newblock Edge device detection of tea leaves with one bud and two leaves based
  on shufflenetv2-yolov5-lite-e.
\newblock {\em Agronomy}, 13(2):577, 2023.

\end{thebibliography}

\end{document}